\documentclass[10pt,twocolumn,letterpaper]{article}

\usepackage{iccv}
\usepackage{times}
\usepackage{epsfig}
\usepackage{graphicx}
\usepackage{amsmath}
\usepackage{amssymb}

\usepackage{color}

\usepackage{multirow}
\usepackage{stfloats}

\usepackage{amsthm}
\usepackage{comment}
\usepackage{listings}
\usepackage{algorithm}

\usepackage{siunitx}
\usepackage{cite}
\usepackage{ctable}
\usepackage{hhline}
\usepackage{booktabs}
\usepackage{subcaption}
\usepackage{csquotes}
\usepackage{adjustbox}
\usepackage{tabu}
\usepackage{enumitem}
\usepackage{tabularx}
\usepackage[normalem]{ulem}
\usepackage{multirow}
\usepackage{colortbl}
\usepackage{booktabs}
\usepackage{tablefootnote}
\usepackage{wrapfig}

\definecolor{LightCyan}{rgb}{0.9059,0.9961,1}
\usepackage{multirow}
\usepackage{graphicx}
\usepackage[font=footnotesize,labelfont=bf,aboveskip=1pt,belowskip=-11pt]{caption}  
\usepackage{makecell}
\definecolor{demphcolor}{RGB}{144,144,144}

\usepackage{pifont}
\newlength\savewidth

\newcommand{\tablestyle}[2]{\setlength{\tabcolsep}{#1}\renewcommand{\arraystretch}{#2}\centering\footnotesize}
\makeatletter\renewcommand\paragraph{\@startsection{paragraph}{4}{\z@}
  {.5em \@plus1ex \@minus.2ex}{-.5em}{\normalfont\normalsize\bfseries}}\makeatother
\newdimen\abovecrulesep
\newdimen\belowcrulesep
\makeatletter
\patchcmd{\@@@cmidrule}{\aboverulesep}{\abovecrulesep}{}{}
\patchcmd{\@xcmidrule}{\belowrulesep}{\belowcrulesep}{}{}
\makeatother
\usepackage{etoolbox}
\preto\align{\par\nobreak\small\noindent}
\expandafter\preto\csname align*\endcsname{\nobreak\small}
\preto\equation{\small}

\usepackage[pagebackref=true,breaklinks=true,letterpaper=true,colorlinks,bookmarks=false]{hyperref}

\iccvfinalcopy 
\usepackage{color, colortbl}
\definecolor{Gray}{gray}{0.9}
\definecolor{Red}{RGB}{230, 57, 70}
\def\iccvPaperID{6690} 

\newcommand{\cmark}{\ding{51}}%
\newcommand{\xmark}{\ding{55}}%

\newcommand{\ourmodel}{\textit{OpenSeeD}}

\ificcvfinal\fi

\begin{document}

\title{A Simple Framework for Open-Vocabulary Segmentation and Detection}

\author{\textbf{ ~Hao Zhang$^{1,2}$\thanks{Equal contribution. List in random.}, Feng Li$^{1,2*}$, ~Xueyan Zou$^{4}$, ~Shilong Liu$^{2,5}$, ~Chunyuan Li$^{3}$} \\ \textbf{~Jianfeng Gao$^{3}$, ~Jianwei Yang$^{3}$\thanks{Equal advisory contribution.}, ~Lei Zhang$^{2\dag}$\hspace{1.mm}} \\
$^1$\small{The Hong Kong University of Science and Technology.} \\
$^2$\small{International Digital Economy Academy (IDEA).} \\
$^3$\small{Microsoft Research at Redmond.} 
$^4$\small{University of Wisconsin-Madison.} \\
$^5$\small{Dept. of CST., BNRist Center, Institute for AI, Tsinghua University.}
\\
\centerline{\tt\tiny  
\{hzhangcx, fliay\}@connect.ust.hk
 \ \{xueyan\}@cs.wisc.edu 
 \ \{liusl20\}@mails.tsinghua.edu.cn
 \ \{jianwyan,jfgao,chunyl\}@microsoft.com
 \ \{leizhang\}@idea.edu.cn
}
}
\twocolumn[{%
\renewcommand\twocolumn[1][]{#1}%
\maketitle
\vspace{-2.5em}
\begin{center}
    \centering
    \includegraphics[width=0.98\textwidth]{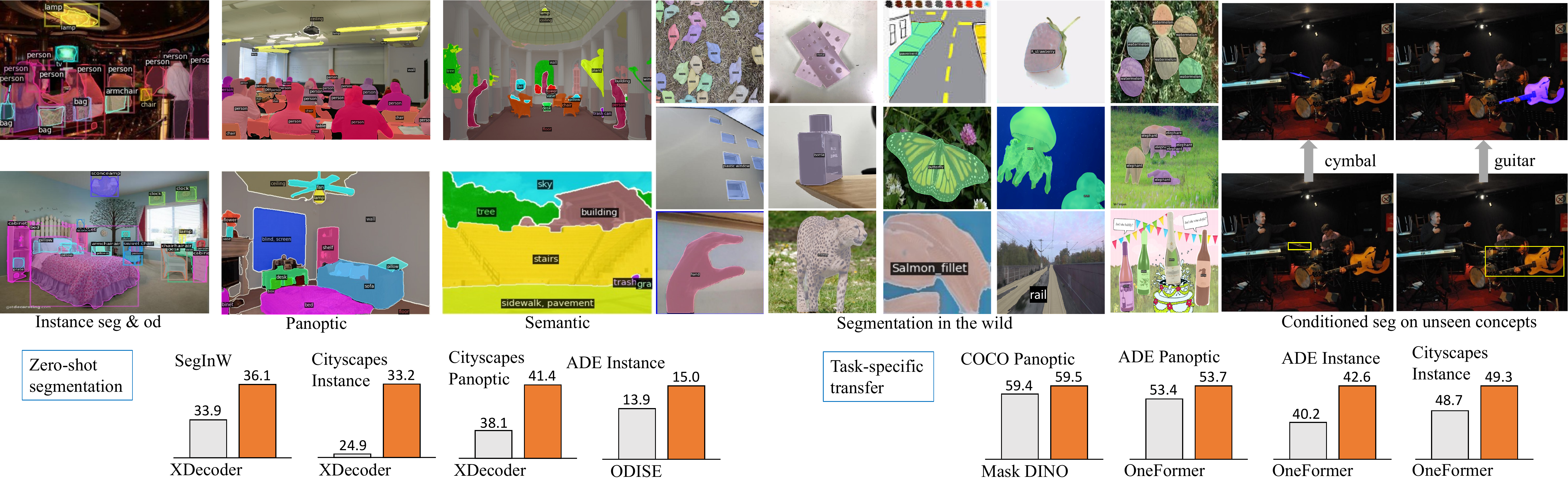}  
\vspace{0.2em}
\captionof{figure}{Upper row: visualizations of \ourmodel{} for open-vocabulary instance segmentation and detection, panoptic and semantic segmentation, instance segmentation in the wild, and conditioned segmentation given referring box location and concept. Lower row: Our \ourmodel{} model outperforms previous state-of-the-art methods (listed below each gray bar) on eight benchmarks in zero-shot and task-specific transfer settings.
}
\label{fig:intro}
\vspace{1.0em}
\end{center}%
}]

\maketitle
\ificcvfinal\thispagestyle{empty}\fi

\begin{abstract}
\let\thefootnote\relax\footnotetext{$^*$Equal contribution. List in random.}
\let\thefootnote\relax\footnotetext{$^\dag$Equal advisory contribution.}
   We present \ourmodel{}, a simple \textbf{Open}-vocabulary \textbf{Se}gm\textbf{e}ntation and \textbf{D}etection framework that jointly learns from different segmentation and detection datasets. 
   To bridge the gap of vocabulary and annotation granularity, we first introduce a pre-trained text encoder to encode all the visual concepts in two tasks and learn a common semantic space for them. This gives us reasonably good results compared with the counterparts trained on segmentation task only. To further reconcile them, we locate two discrepancies: $i$) task discrepancy -- segmentation requires extracting masks for both foreground objects and background stuff, while detection merely cares about the former; $ii$) data discrepancy -- box and mask annotations are with different spatial granularity, and thus not directly interchangeable. To address these issues, we propose a decoupled decoding to reduce the interference between foreground/background and a conditioned mask decoding to assist in generating masks for given boxes. To this end, we develop a simple encoder-decoder model encompassing all three techniques and train it jointly on COCO and \let\thefootnote\relax\footnotetext{This work is developed during an internship at IDEA.}
   \noindent Objects365. After pre-training, our model exhibits competitive or stronger zero-shot transferability for both segmentation and detection. Specifically, \ourmodel{} beats the state-of-the-art method for open-vocabulary instance and panoptic segmentation across 5 datasets, and outperforms previous work for open-vocabulary detection on LVIS and ODinW under similar settings. When transferred to specific tasks, our model achieves new SoTA for panoptic segmentation on COCO and ADE20K, and instance segmentation on ADE20K and Cityscapes. The lower row in Fig.~\ref{fig:intro} shows a comparison of the performance of \ourmodel{} and previous SoTA methods. Finally, we note that \ourmodel{} is the first to explore the potential of joint training on segmentation and detection, and hope it can be received as a strong baseline for developing a single model for both tasks in open world. Code will be released at \url{https://github.com/IDEA-Research/OpenSeeD}.
   
\end{abstract}
\section{Introduction}
Developing vision systems that can be transferable to novel concepts or domains has emerged as an important research topic in the community. In the light of strong zero-shot transferability demonstrated in the seminal work CLIP~\cite{radford2021learning}, a number of researchers have attempted to build advanced open-vocabulary models by leveraging large-scale image-text pairs for fine-grained vision tasks like detection~\cite{gu2021open,li2022grounded,zhong2022regionclip,gao2022open,kuo2022f} and segmentation~\cite{ding2022open,huynh2022open,xu2023side,zhou2021denseclip}.

\begin{figure}[t]
    \centering
    \includegraphics[width=0.95\linewidth]
    {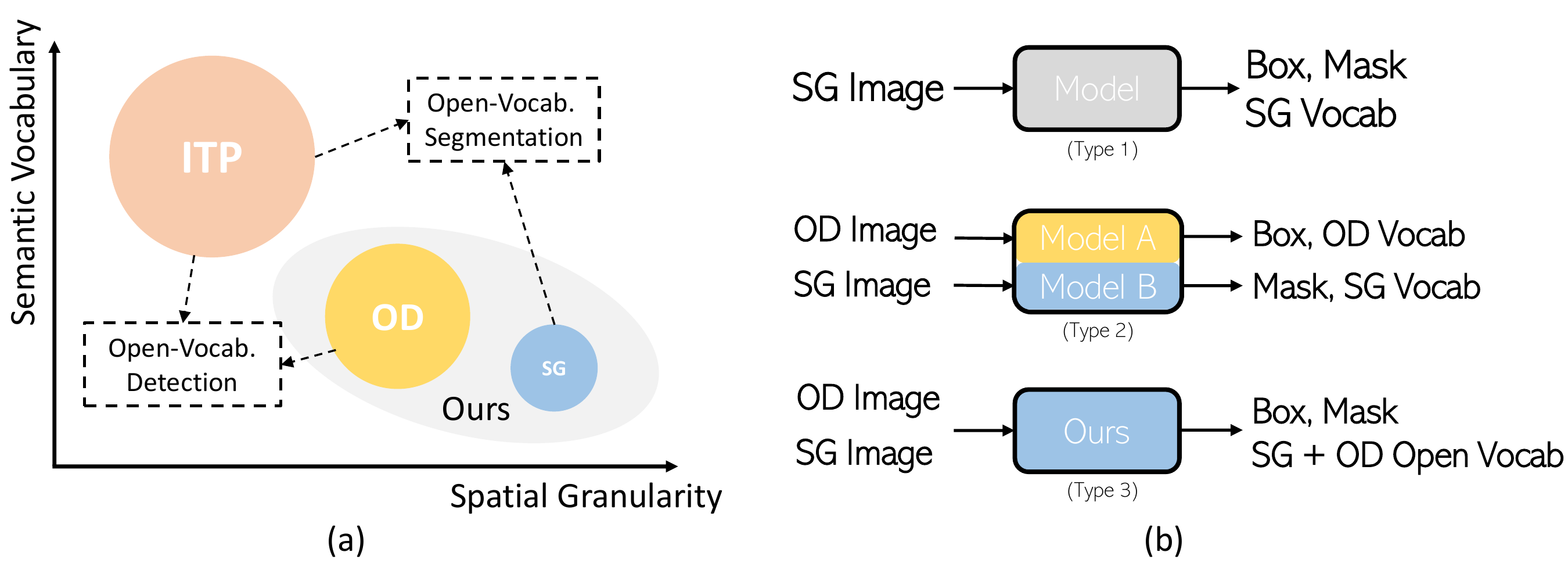}
    \caption{(a) Semantic vocabulary sizes and spatial granularities comparison across different vision tasks/datasets. ``ITP'' means image-text pairs; ``OD'' means object detection and ``SG'' means segmentation. Our \ourmodel{} is the first open-vocabulary model that jointly learn on segmentation and detection (gray region). (b) Different types of methods connect object detection and segmentation.}
    \label{fig:teaser}
    \vspace{-3pt}
\end{figure}

Arguably, core vision tasks like detection and segmentation are fairly distinct in their vocabulary sizes and spatial granularities of supervision, as illustrated in Fig.~\ref{fig:teaser}~(a). For example, the commonly used public detection dataset Objects365~\cite{shao2019objects365} contains box annotations for 365 concepts in around 1.7M images, while mask annotations in COCO~\cite{lin2014microsoft} cover merely 133 categories in 0.1M images. Previous works have explored different ways of leveraging a large amount of image-text data for open-vocabulary detection or segmentation, such as distilling the visual-semantic representations from multi-modal foundation models~\cite{gu2021open,zhong2022regionclip}, designing fine-grained or augmented contrastive learning methods~\cite{mukhoti2022open} or utilizing pseudo-labeling techniques ~\cite{li2022grounded,zhang2022glipv2}. To the best of our knowledge, most (if not all) of them focused on how to improve the performance for either detection or segmentation. Moreover, transferring weak image-level supervision to fine-grained tasks usually requires sophisticated designs to mitigate the huge granularity gap and is vulnerable to noises in image-text pairs. This leads to a natural question: \emph{can we bridge detection and segmentation that are cleaner and have a closer gap to attain a good open-vocabulary model for both?}

Taking one step back, marrying detection and segmentation had been previously explored in two main ways. On one hand, Mask R-CNN~\cite{he2017mask} is one of the first works that proposed to jointly learn detection and instance segmentation on COCO. On the other hand, it is shown that detection models pre-trained on Objects365 can be feasibly transferred for COCO panoptic segmentation~\cite{li2022mask}. However, as depicted in Fig.~\ref{fig:teaser}~(b), the former method requires the model to be trained on the same dataset containing aligned box and mask annotations, while the latter method follows pre-train-then-fine-tune protocol, leading to two separate closed-set models. In this work, we are the first to propose jointly learning from detection and segmentation data, and more importantly serving an open-vocabulary model for both tasks (Fig.~\ref{fig:teaser}~(b) bottom). 
Achieving this goal requires answering two critical questions: $i$) how to transfer the semantic knowledge across detection and segmentation data; $ii$) how to bridge the gap between box and mask supervision. First, the vocabulary shares commons but also bear substantial differences between the two tasks. We need to accommodate the two vocabularies and further go beyond towards \textit{open} vocabulary. Second, semantic and panoptic segmentation tasks require segmenting not only foreground objects (things like ``dog'' and ``cat''.) but also background concepts (stuff like ``sky'' and ``building''), while detection task solely cares about foreground objects. Third, box supervision by nature is coarser than mask supervision. We can convert masks into boxes but hardly vice versa.

To the end, we propose \ourmodel{}, a simple encoder-decoder framework to reconcile the two tasks by mitigating the aforementioned problems. Concretely, we first exploit a single text encoder to encode all concepts occurring in the data and train our model to align the visual tokens with the semantics in a common space. Second, we explicitly divide the object queries in the decoder into two sub-types: foreground and background queries, where the first group is responsible for foreground objects from both segmentation and detection while the second group is only for background stuffs in segmentation. Third, we introduce conditioned mask decoding which learns to decode masks from ground-truth boxes from segmentation data and generates the mask assistant for detection data. 
As a result, our \ourmodel{} is able to learn from separate detection and segmentation data seamlessly and achieves outstanding or competitive zero-shot and transfer performance across various tasks/datasets. Fig~\ref{fig:intro} shows a visualization of our model on instance, panoptic and semantic segmentation tasks. It also shows the segmentation results on datasets that largely differ from our training data such as the SeginW datasets and demonstrates the conditioned segmentation ability of \ourmodel{}. \textit{Given the encouraging results, we hope our work can contribute as the first strong baseline for developing a single open-vocabulary model for both tasks}.

\noindent
\textbf{Contributions}. To summarize, our main contributions are:
\begin{itemize}[nolistsep]
    \item We are the first to present a strong baseline model that can jointly learn from detection and segmentation data towards an open-vocabulary model for both tasks.
    \item We locate the discrepancies in two tasks/datasets and propose separate techniques including shared semantic space, decoupled decoding, and conditioned mask assistance to mitigate the issues.
    \item By jointly training our model on segmentation and detection data, we achieve new state-of-the-art segmentation performance for zero-shot and task transfer across a variety of datasets, and competitive performance for zero-shot object detection.
\end{itemize}

\section{Related Work}
\noindent
\textbf{Generic Segmentation and Detection.}
Detection and segmentation have been long-standing problems in the vision community~\cite{fu1981survey,felzenszwalb2009object,zou2019object,minaee2021image}. Both tasks require understanding what and where the visual concepts are but with different spatial granularities. Generic segmentation mainly includes instance, semantic and panoptic segmentation~\cite{he2017mask,chen2017deeplab,kirillov2019panoptic}, with respect to different semantics. Recently, Detection Transformer (DETR)~\cite{carion2020end} that is based on Transformer~\cite{vaswani2017attention} has achieved significant progress in many detection~\cite{zhu2020deformable, liu2022dabdetr, meng2021conditional, li2022dn, zhang2022dino} and segmentation models~\cite{li2022panoptic, cheng2022masked, li2022mask, jain2022oneformer}. However, all these methods are constrained to a limited vocabulary size.
\\
\noindent
\textbf{Open-Vocabulary Segmentation. } Many open-vocabulary segmentation models~\cite{li2022language, ghiasi2021open, huynh2022open, ding2022open} leverages large pretrained vision-language models (\textit{e.g.}, CLIP~\cite{jia2021scaling} or ALIGN\cite{radford2021learning}) to distill or transfer the visual-semantic knowledge. Apart from using foundation models, DenseCLIP~\cite{rao2022denseclip} and GroupViT~\cite{xu2022groupvit} show that fine-tuning from a foundation model or training from scratch can also yield superior zero-shot performance. Recently, X-Decoder~\cite{zou2022generalized} proposes to unify all types of segmentation tasks and several vision-language tasks for open-vocabulary segmentation. In ODISE~\cite{xu2023open}, the authors study a new way of using a text-to-image diffusion model as the backbone for open-vocabulary segmentation. Unlike the previous works, our model instead explores connecting segmentation and detection which have cleaner data and closer gap between each other.
\\
\noindent
\textbf{Open-Vocabulary Detection}. Similarly, some open-vocabulary detection models directly leverage foundation models for distillation or transfer like OV-DETR~\cite{zareian2021open} and VILD~\cite{XiuyeGu2021OpenvocabularyOD}. Recently, GLIP~\cite{li2022grounded} formulates detection as a special grounding problem to unify detection and phrase grounding tasks. These grounding data help improve the alignment between phrases and regions for open detection. RegionCLIP~\cite{zhong2022regionclip} and DetCLIP~\cite{LeweiYao2022DetCLIPDV} generate pseudo box labels from image-text pairs for more generalized detection. 
\\
\noindent
\textbf{Weakly-Supervised Segmentation}. Weakly-supervised segmentation typically only uses box annotation as supervision to generate segmentation. Prominent methods design teacher models or weak supervision loss, like BoxInst~\cite{tian2021boxinst},  Box2Mask~\cite{li2022box}, DiscoBox~\cite{lan2021discobox} and Mask Auto-Labelers~\cite{lan2023vision}. All these models are with closed-set and usually inferior to models with segmentation supervision. In contrast, we attempt to leverage as much supervision as possible from both segmentation and detection for an open-vocabulary model.
\\
\noindent
\textbf{Learning from Box and Mask}. There are primarily two ways to learn from both box and mask. The first one is to train on a single dataset with both box and mask annotations. Prominent methods include Mask R-CNN~\cite{he2017mask} and HTC~\cite{chen2019hybrid}. However, they are constrained to foreground instances. The second way is to pretrain with only box supervision and then transfer to segmentation. For example, HTC~\cite{chen2019hybrid} and Mask DINO~\cite{li2022mask} can both learn from large-scale detection data and then be fine-tuned to a specific segmentation dataset. However, such a pretrain-and-finetune protocol leads to two separate models that are only capable of either detection or segmentation. Moreover, both models are closed-set and thus not transferable to novel concepts.




\section{Method}
\begin{figure*}[t]
    \centering
    \includegraphics[width=0.95\linewidth]{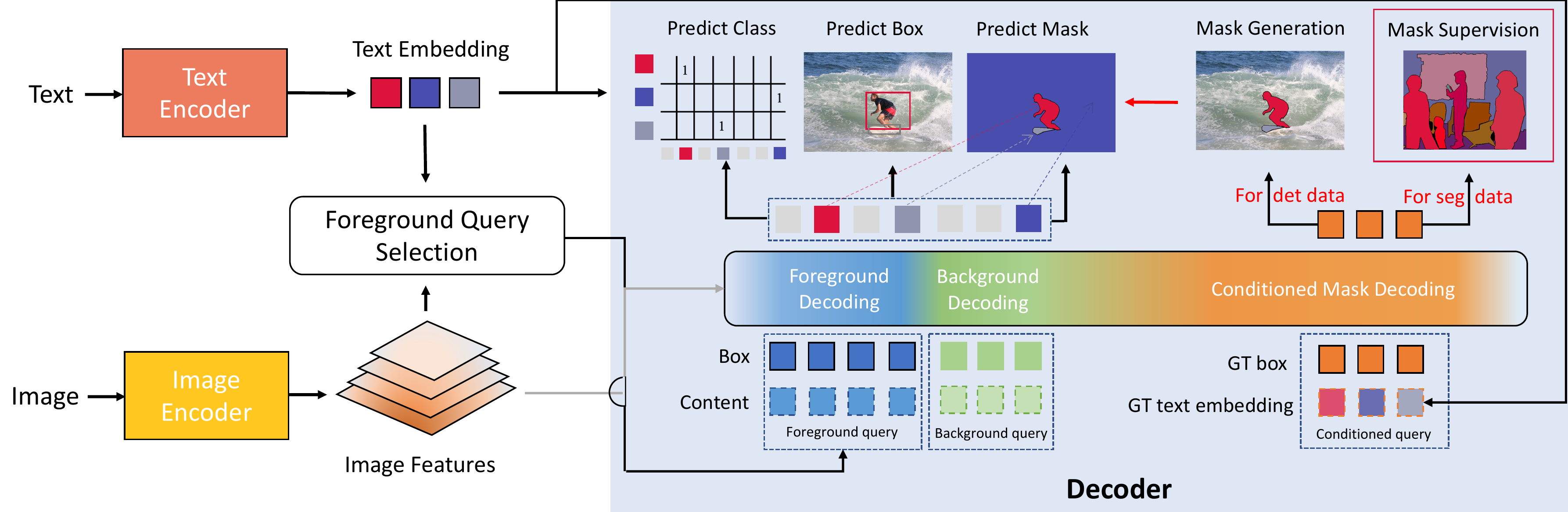}
    \caption{The framework of \ourmodel{}. The framework consists of one image encoder, one text encoder, and our designed decoder with foreground, background, and conditioned mask decoding capability.  ``GT'' means ground truth, and ``GT text embedding'' is the embedding encoded by the text encoder. 
    }
    \label{fig:framework}
\end{figure*}

    


Given segmentation and detection datasets, \ourmodel{} is aimed at learning an open-vocabulary model for both tasks. Formally, let $\mathcal{D}_m = \{{I}_i, (\mathbf{c}_i, \mathbf{m}_i)\}_{i=1}^{M}$ denote the segmentation dataset of size $M$ and $\mathcal{D}_b = \{I_j, (\mathbf{c}_j, \mathbf{b}_j)\}_{j=1}^{N}$ the detection dataset of size $N$, where $\mathbf{c}$ are the visual concepts in an image, and $\mathbf{m}$ and $\mathbf{b}$ the corresponding masks and boxes, respectively. Suppose $\mathcal{V}=\{c_1,...c_K\}$ be the vocabulary of unique $K$ visual concepts appearing in $\mathcal{D}_m$ and $\mathcal{D}_b$. The goal of \ourmodel{} is learning to detect \textit{and} segment visual concepts in $\mathcal{V}$ and beyond.

To achieve the goal, we exploit a general encoder-decoder design and employ a text encoder for our \ourmodel{}, as shown in Fig.~\ref{fig:framework}. Our model takes as input an image $I$ and the vocabulary $\mathcal{V}$ and output a set of predictions including masks $\mathbf{P^m}$, boxes $\mathbf{P^b}$, and classification scores $\mathbf{P^c}$. As a whole, $\langle\mathbf{P^m}, \mathbf{P^b}, \mathbf{P^c}\rangle=\mathsf{OpenSeeD}(I,\mathcal{V})$. More specifically, our model consists of one image encoder $\mathsf{Enc_I}$, one text encoder $\mathsf{Enc_T}$, and one decoder $\mathsf{Dec}$. Given an image $I$ and the vocabulary $\mathcal{V}$, we first encode them by $\mathsf{Enc_I}$ and $\mathsf{Enc_T}$, respectively: 
\begin{equation}
    \mathbf{O}=\mathsf{Enc_I}(I), \mathbf{T}=\mathsf{Enc_T}(\mathcal{V})
\end{equation}
where the image features $\mathbf{O}\in \mathcal{R}^{H \times W \times C}$, 
and the text features $\mathbf{T}= \{t_1, t_2, ..., t_K\}$. Afterward, the decoder takes $L$ queries $\mathbf{Q} \in \mathcal{R}^{L \times C}$ as inputs and cross-attends the image features to get outputs:
\begin{equation}
\begin{aligned}
    \langle\mathbf{P^m}, \mathbf{P^b}, \mathbf{P^s}\rangle &= \mathsf{Dec}\left(\mathbf{Q}; \mathbf{O}\right) \\
    \mathbf{P^c}&=\mathsf{Sim}(\mathbf{P^s}, \mathbf{T})
    \end{aligned}
    \label{EQ:uniseg_overall}
\end{equation}
where $\mathbf{P^s}$ is the decoded semantics. The visual-semantic matching scores $\mathbf{P^c}$ is derived from $\mathbf{Sim}(\mathbf{P^s}, \mathbf{T})$ by calculating the similarity scores between $\mathbf{P^s}$ and $\mathbf{T}$, which is used to compute the loss during training and predict the category during inference.

\subsection{Basic Loss Formulation}
In this basic formula, we attempt to reconcile the two tasks by promoting a shared visual-semantic space without touching other issues. For multiple tasks and datasets, our loss function can be written as follows.
\begin{equation}
\footnotesize
\begin{aligned}
    \mathcal{L}_{all}&=\sum_{{I}, (\mathbf{c}, \mathbf{m})   \in \mathcal{D}_m} 
    \overbrace{\left(\mathcal{L}_{m}(\mathbf{P^m},\mathbf{m})+
    \mathcal{L}_{b}(\mathbf{P^b},\hat{\mathbf{b}})+
    \mathcal{L}_{c}(\mathbf{P^c},\mathbf{c})\right)}^{\substack{\text{Segmentation loss}}}\\
    &+\sum_{{I}, (\mathbf{c}, \mathbf{b}) \in \mathcal{D}_b}
    \underbrace{\left(\mathcal{L}_{b}(\mathbf{P^b},\mathbf{b})+
    \mathcal{L}_{c}(\mathbf{P^c},\mathbf{c})\right)}_{\substack{\text{Detection loss}}}\\
\end{aligned}
\label{Eq:base}
\end{equation}
For clarity, we omit the weight for each loss term. Note that for the segmentation task, we can derive accurate boxes $\hat{\mathbf{b}}$ from masks $\mathbf{m}$ and use them to compute the box loss as in term $\mathcal{L}_{b}(\mathbf{P^b},\hat{\mathbf{b}})$.
By summing over all the terms, our model can achieve a reasonably good open-vocabulary performance. Furthermore, it can be pre-trained end-to-end with detection and segmentation data, allowing it to perform open-vocabulary segmentation and detection using a single set of weights.

Despite building a strong baseline, we must consider the intrinsic discrepancies between the two tasks, as previously discussed. Semantic and panoptic segmentation require the recognition of both foreground and background, while detection focuses solely on localizing foreground objects. As a result, using the same queries for both tasks creates conflicts that can significantly degrade performance. Additionally, good box predictions are typically indicative of good masks, and vice versa. Separately training the box and mask head on detection and segmentation data obstructs the synergy of spatial supervision from both datasets. 

To address the aforementioned discrepancies, we introduce a new decoder design for our \ourmodel{}. We divide the queries $\mathbf{Q}$ into three types: $L_f$ foreground queries $\mathbf{Q_f}$, $L_b$ background queries $\mathbf{Q_b}$ and $L_d$ conditioned queries $\mathbf{Q_d}$, and propose query-specific computations for each type. In the following, we will describe how we decouple the foreground and background decoding to address the task discrepancy in Sec.~\ref{sec:task_discrepancy}, and employ the conditioned mask decoding to tackle the data discrepancy in Sec.~\ref{sec:data_discrepancy}. 
\subsection{Bridge Task Gap: Decoupled Foreground and Background Decoding}\label{sec:task_discrepancy}
\begin{figure}[t]
    \centering
    \includegraphics[width=0.95\linewidth]
    {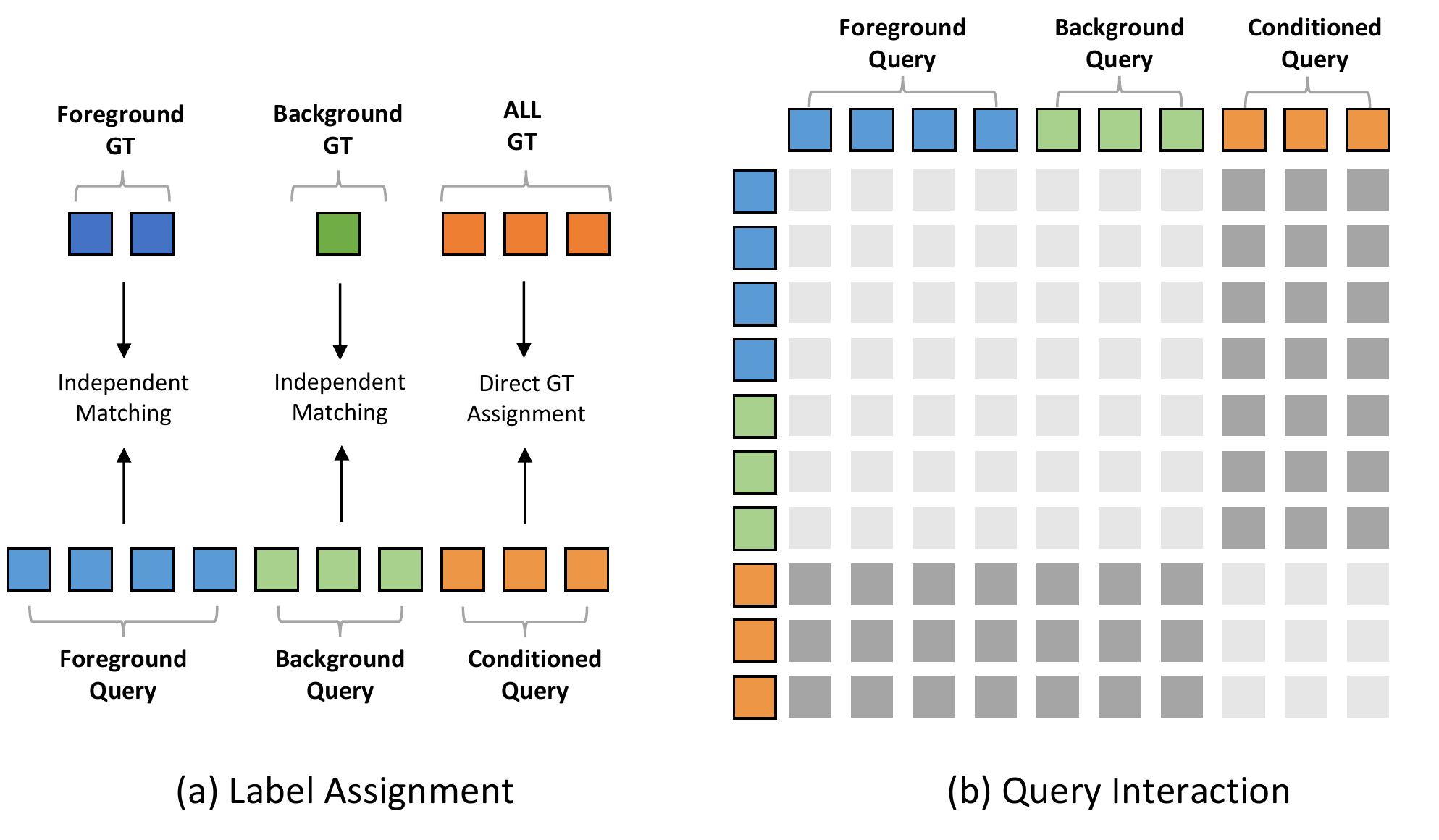}
    \caption{a) Label assignment for different queries. ``gray'' means unused for detection data without background stuff.  Best viewed in color. b) Query interaction of different queries (foreground, background, and conditioned query) in \ourmodel{}. All data has background queries including detection data.  ``dark'' color means blocking interaction. }
    \label{fig:interaction_assignment}
    \vspace{-3pt}
\end{figure}
Without loss of generality, we have defined the visual concepts that appear in instance segmentation and detection as foreground, while the stuff categories in panoptic segmentation are considered background. To mitigate the task discrepancy, we perform foreground and background decoding with foreground queries $\mathbf{Q_f}$ and background queries $\mathbf{Q_b}$, respectively. Specifically, for these two query types, our decoder predicts two sets of outputs: $\langle\mathbf{P}^{m}_f, \mathbf{P}^{b}_f, \mathbf{P}^c_f \rangle$ and $\langle\mathbf{P}^{m}_b, \mathbf{P}^{b}_b, \mathbf{P}^c_b \rangle$. We also divide the ground truths in segmentation dataset into two groups: $(\mathbf{c}_f, \mathbf{m}_f)$ and $(\mathbf{c}_b, \mathbf{m}_b)$, and then perform two independent Hungarian Matching processes for these two sets correspondingly, as shown in Fig.\ref{fig:interaction_assignment}~(a). Consequently, both foreground and background decoding are used for segmentation, while only foreground decoding is used for detection. As a result, our basic loss function in Eq.~\eqref{Eq:base} is reformulated to:
\begin{equation}
\footnotesize
\begin{aligned}
    \mathcal{L}_{all}&=\sum_{{I}, (\mathbf{c}, \mathbf{m}) \in \mathcal{D}_m} \overbrace{\left(\mathcal{L}_{m}(\mathbf{P^m_f},\mathbf{m}_f)+
    \mathcal{L}_{b}(\mathbf{P^b_f},\hat{\mathbf{b}}_f)+
    \mathcal{L}_{c}(\mathbf{P^c_f},\mathbf{c}_f)\right)}^{\substack{\text{Segmentation loss for foreground}}} \\
    &+\overbrace{\left(\mathcal{L}_{m}(\mathbf{P^m_b},\mathbf{m}_b)+L_{b}(\mathbf{P^b_b},\hat{\mathbf{b}}_b)+
    \mathcal{L}_{c}(\mathbf{P^c_b},\mathbf{c}_b)\right)}^{\substack{\text{Segmentation loss for background}}} \\
    &+\sum_{{I}, (\mathbf{c}, \mathbf{b}) \in \mathcal{D}_b} \underbrace{\left(\mathcal{L}_{b}({\mathbf{P^b_f}},\mathbf{b})+
    \mathcal{L}_{c}({\mathbf{P^c_f}},\mathbf{c})\right)}_{\substack{\text{Detection loss for foreground}}}\\
\end{aligned}
\label{decouple}
\end{equation}
where $\hat{\mathbf{b}}_f$ and $\hat{\mathbf{b}}_b$ are derived from ${\mathbf{m}}_f$ and ${\mathbf{m}}_b$, respectively. Based on such explicit decoupling, our model maximizes the cooperation of foreground supervision from both detection and segmentation datasets and significantly reduces the interference between foreground and background categories. Though decoupled, we note that these two types of queries share the same decoder and interact with each other with self-attention, as shown in Fig.~\ref{fig:interaction_assignment}~(b). 
Below we explain how the foreground and background queries are determined.


\noindent
\textbf{Language-guided foreground query selection}. Open-vocabulary setting differs from the conventional closed-set setting in that a model is required to localize a large number of foreground objects far beyond the training vocabulary. However, the fact is that our decoder contains a limited number of foreground queries (a few hundred typically), making it hardly handle all possible concepts in the image. 
\begin{figure}[t]
    \centering
    \includegraphics[width=0.95\linewidth]
    {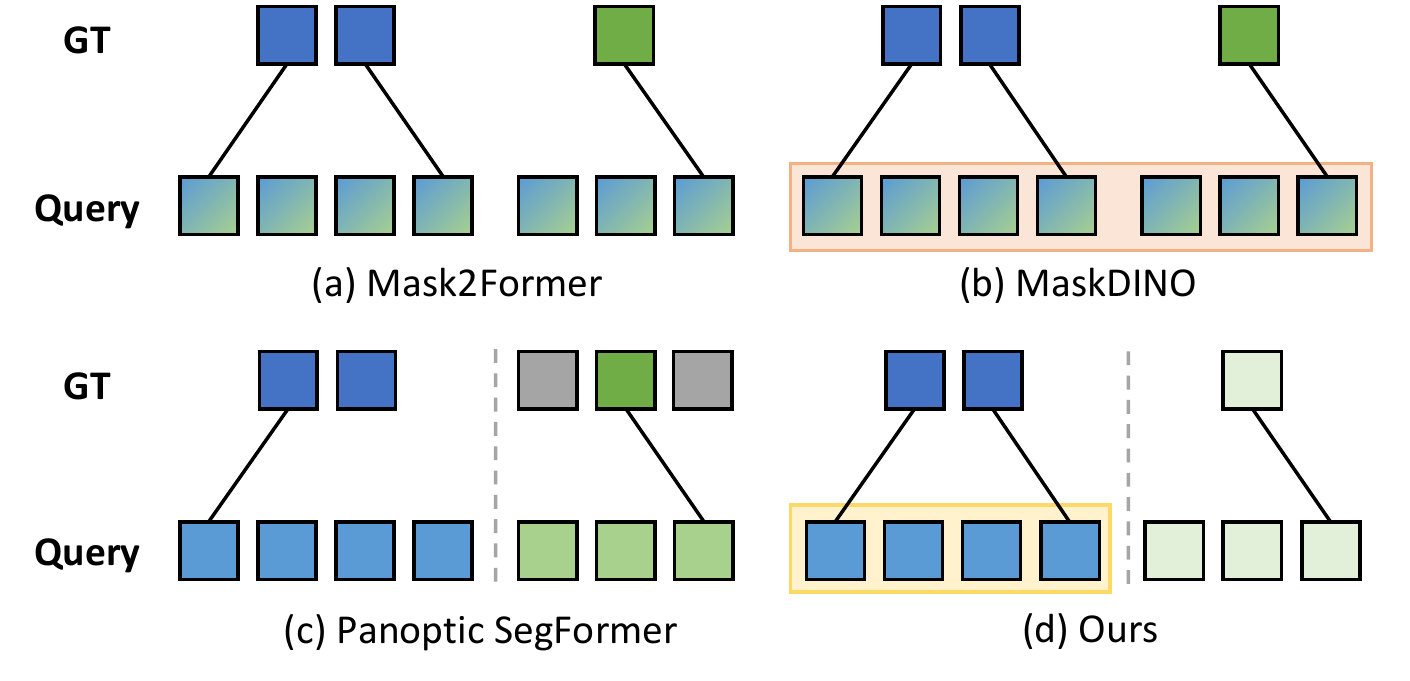}
    \caption{Compare different ways of handling foreground (fg) and background (bg) queries. 
    Note: blue: fg. green: bg. mixed bule\&green: treat fg and bg equally. Shaded queries are acquired by query selection from image features, others are learnable. a)\&b) Mask2Former and MaskDINO teat fg and bg equally. They differ in that queries are learnable for Mask2Former while selected for MaskDINO. c) PanopticSegFormer discriminates fg and bg, but assign each bg query to a fixed category. d) Ours. The bg queries are lightgreen as they are not matched to GT in detection data.}
    \label{fig:match compare}
    \vspace{0.8cm}
\end{figure}
To address this issue, we propose a method called \textit{language-guided foreground query selection} to adaptively select queries with respect to given text concepts as shown in Fig.~\ref{fig:framework} left part. 
Given the image features $\mathbf{O}$ and text features $\mathbf{T}$, we employ a lightweight module to predict the box and score for each feature: 
\begin{equation}
    \mathbf{E}^b = \mathsf{Head}\left(\mathbf{O}\right),    \mathbf{E}^c =\mathsf{Sim}(\mathbf{O}, \mathbf{T}) \\
    \label{EQ:uniseg_overall}
\end{equation}
where $\mathsf{Head}$ is the box head. Then we select $L_f$ top-ranked entries from $\mathbf{E}^b$ and $\mathbf{O}$ according to the scores in $\mathbf{E}^c$.
These selected $L_f$ image features and boxes are then fed to the decoder as the foreground queries (blue squares in Fig.~\ref{fig:framework}). By selecting only the text-related tokens as decoder queries, we mitigate the problem of decoding irrelevant semantics and provide better query initialization. Such an adaptive way of proposing foreground queries enables our model to effectively transfer to novel vocabulary during test scenarios.

\noindent
\textbf{Learnable background queries}. Different from foreground queries, we use learnable query embeddings for our background queries for two reasons. Firstly, query selection does not work well because the selected reference points often extend beyond large and non-convex background regions, leading to suboptimal results. Secondly, background stuff has a relatively smaller number of categories than the foreground, and a single image typically contains a few different stuffs (\textit{e.g.}, ``sky'', ``building''). As a result, using learnable queries for our model can sufficiently and effectively handle background stuff categories and generalize well to open-vocabulary settings. The background queries are marked by green squares in Fig.~\ref{fig:framework}.
\begin{figure}[t]
    \centering
    \vspace{4pt}
    \includegraphics[width=0.99\linewidth]{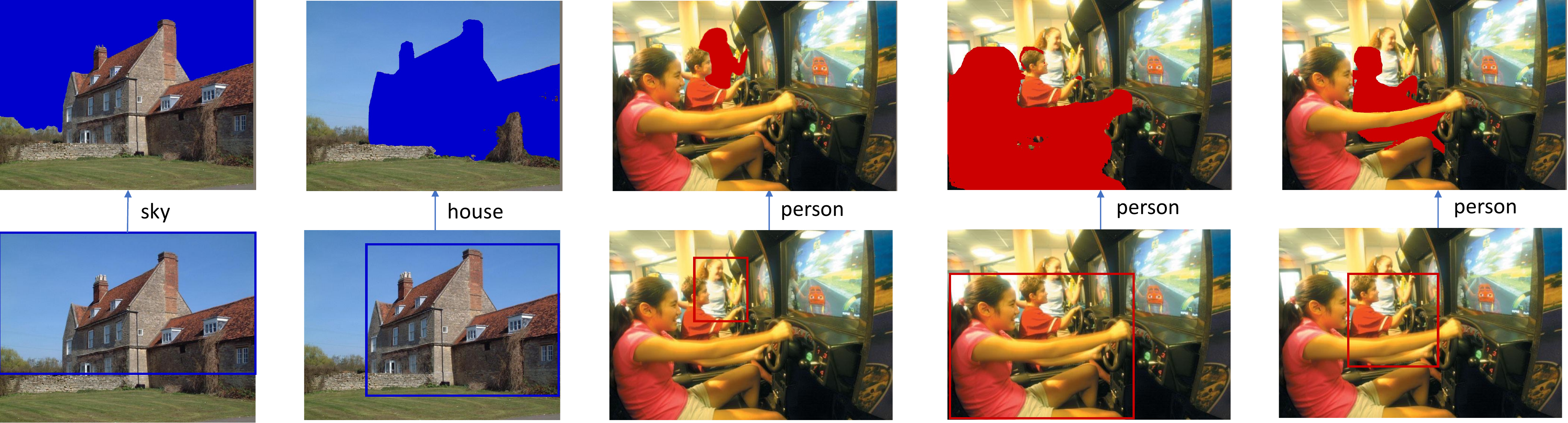}
    \caption{Background and foreground masks generated with boxes and text as the condition for sample images in ADE20K~\cite{zhou2017scene}. }
    \label{fig:decoded_mask}
\end{figure}
\\
\textbf{Comparison with previous works}. 
In Fig.~\ref{fig:match compare}, we show a comparison between our approach and others on handling foreground and background. 
Mask2Former\cite{cheng2022masked} and MaskDINO~\cite{li2022mask} treat foreground and background equally when conducting panoptic segmentation, resulting in suboptimal mask average precision (AP) for foreground objects compared to the same model solely trained on foreground classes (instance segmentation). Panoptic Segformer~\cite{li2022panoptic} separates foreground and background queries, but their background queries have fixed semantics, with each query corresponding to a pre-defined background category, limiting their ability to handle open-vocabulary categories. In contrast, our approach proposes foreground queries through a language-guided selection mechanism, and our background queries are fully learnable, eliminating the restrictions of a predefined vocabulary.
\subsection{Bridge Data Gap: Conditioned Mask Decoding}
\label{sec:data_discrepancy}
Our ultimate goal is to bridge the data gap by using a single loss function to train multiple tasks, resulting in the following loss function:
\begin{equation}
\footnotesize
\begin{aligned}
    \mathcal{L}_{all}&=\sum_{{I}, (\mathbf{c}, \mathbf{m}, \mathbf{b})   \in \mathcal{D}} 
    \mathcal{L}_{m}(\mathbf{P^m},\mathbf{m})+
    \mathcal{L}_{b}(\mathbf{P^b},\mathbf{b})+
    \mathcal{L}_{c}(\mathbf{P^c},\mathbf{c})
\end{aligned}
\label{Eq:unified}
\end{equation}
Here, $\mathcal{D}$ represents the union of segmentation and detection datasets. However, the loss function requires mask annotations for detection data and box annotations for segmentation data, leading to a discrepancy in the granularity of spatial supervision between the two tasks. 
As we discussed earlier, we can easily convert an object mask $m$ to a box $\hat{b}$, which augments the original segmentation data $\mathcal{D}_m=\{I_i, (\mathbf{c}_i, \mathbf{m}_i)\}_{i=1}^{M}$ into $\hat{\mathcal{D}}_m=\{I_i, (\mathbf{c}_i, \mathbf{m}_i, \hat{\mathbf{b}}_i)\}_{i=1}^{M}$. For detection data $\mathcal{D}_b$, however, we are only given coarse location (box) and category. Then an interesting question comes -- \textit{can we obtain its mask given these priors?} 
\begin{table}[t]
\vspace{0.5cm}
\caption{Results for models trained on COCO without and with conditioned mask decoding. Models are evaluated on COCO and ADE20K validation set. ``final'' and ``early'' means the model at the final and early training stage, respectively. ``Convert box into mask'' means we directly convert the GT boxes into rectangle masks for evaluation.}
\label{tab:task_transfer}
\begin{adjustbox}{width=0.45\textwidth,center}
\begin{tabular}{l|c|c} 
\toprule
\multirow{2}{*}{Method}                        
&{COCO}   & \multicolumn{1}{c}{ADE} \\
& Mask AP    & Mask AP  \\ 
\hline
        \ourmodel{} (T) & 45.1&8.6\\
        \ourmodel{} (T) w conditioned train \& eval~(final)&53.2&\textbf{46.4}\\
        \ourmodel{} (T) w conditioned train \& eval~(early)&42.2&14.8\\
        Convert box into mask&16.2&25.4\\
\bottomrule
\end{tabular}
\end{adjustbox}
\vspace{0.2cm}
\label{tab:box2mask}
\end{table}

To address this problem, we resort to the segmentation data which contains rich mappings from label\&box to mask, \textit{i.e.}, $(c, b) \rightarrow m$ and propose conditioned mask decoding to learn the mappings as shown in Fig.~\ref{fig:framework} right-most part. Given the ground-truth concepts and boxes, $(\mathbf{c}, \mathbf{b})$, we employ the decoder to decode the mask:
\begin{equation}
    \mathbf{P^m} = \mathsf{Dec}\left((\mathbf{t}, \mathbf{b}); \mathbf{O}\right)
    \label{Eq:mask_decoding}
\end{equation}
where $t$ is the text features extracted for the concepts. Based on Eq.~\eqref{Eq:mask_decoding}, the question becomes, \textit{can we learn from segmentation data a good mapping which generalizes well to detection data with different categories?}
\\
\textbf{Mapping Hypothesis Verification}. To answer the question, we conduct a pilot study. 
We train a model which learns to decode masks conditioned on GT concepts and boxes on COCO~\cite{chen2015microsoftcoco}, 
and then evaluate the conditioned decoding performance on ADE20K~\cite{zhou2017scene}. The results are shown in Table~\ref{tab:box2mask}. Comparing the top two rows, we can find mask decoding conditioned on the GT concept and box significantly improves the quality (mask AP from 8.6 to 46.4), which even reaches a similar level to COCO (46.4 \textit{v.s.} 53.2). These results indicate that our learned mask decoding generalizes well to a new dataset with novel categories. To further verify, we visualize decoded masks in Fig.~\ref{fig:decoded_mask}. 

\underline{\textit{Interactive Segmentation}}. 
The above study implies a new interface of image segmentation. Apart from segmenting an image from scratch, users can give a hint about the object location by drawing a box (click four points), and our \ourmodel{} can generate its mask with fairly high quality. This capacity can potentially help accelerate the annotation of segmentation data, especially for those with boxes. 
We leave a comprehensive study on this as future work.
\\
\textbf{Conditioned Mask Decoding Training. }
Based on the verified hypothesis, we add all the GT boxes and labels as the conditioned queries to simultaneously learn foreground/background decoding and conditioned mask decoding, as shown in Fig.~\ref{fig:framework}. 
It unifies all our tasks and enables \ourmodel{} to learn more generalized conditioned decoding in the joint semantic space. Based on this, we can literally derive the pseudo masks $\hat{\mathbf{m}}$ for object detection data and obtain augmented $\hat{\mathcal{D}}_b=\{I_i, (\mathbf{c}_i, \hat{\mathbf{m}_i}, {\mathbf{b}}_i)\}_{i=1}^{N}$. 
Below we elaborate on how these pseudo masks are used for training.

\noindent\textbf{Conditioned Mask Generation to Guide Detection Data. } The trained conditioned mask decoding component can also be used to assist detection data as segmentation guidance. We propose two methods to utilize the generated mask to guide our model training, \textit{Online Mask Assistance} and \textit{Offline Mask Assistance}. For \textit{Online Assistance}, we only train one model and generate the masks on the fly. Instead of directly using the generated masks as mask supervision, we use the masks to assist in matching predictions and GT instances because the mask quality is not strong enough for supervision. especially in the early stage (shown in Tab.~\ref{tab:box2mask} third row). As for \textit{Offline Assistance}, we train our model with conditioned mask decoding until convergence and generate mask annotations for detection data. The annotated dataset can be used to train a segmentation model. Considering detection data only has instance-level annotations, the generated masks are expected to improve instance segmentation in both cases. More details about these two methods are discussed in the Appendix.
\\
\noindent
\textbf{Comparison with Denoising Training}. Compared with models~\cite{li2022dn, zhang2022dino, li2022mask} using denoising training (DN), \textit{conditioned mask decoding} differs in two aspects. First, their design choices are different. DN adds noise to the GT boxes and labels for reconstruction, but our model learns to generate masks conditioned on the GT priors. Second, their design purposes are different. DN is designed to accelerate training convergence~(understanding), while our method aims to generate masks for detection data~(generation).







\begin{table*}
\centering
\tablestyle{3pt}{0.95}
\addtolength{\extrarowheight}{\belowrulesep}
\caption{\textbf{One suit of weights} for open-vocabulary segmentation on multiple datasets in a zero-shot manner. Our model is pre-trained on COCO and Objects365 data. 'SEG' indicates segmentation data (COCO), 'DET' indicates detection data (Objects365), and ITP indicates image-text pairs/referring/captioning data. The values in gray are supervised results. $\star$ X-Decoder (L) is not open-source, so we cannot evaluate its performance on LVIS.}
\label{tab:open_seg}
\footnotesize
\resizebox{0.99\linewidth}{!}{\begin{tabular}{l|ccc|cccc|ccc|cc|cc|cc|c|c} 
\toprule
\multirow{2}{*}{Method} & \multicolumn{3}{c|}{Training Data} & \multicolumn{4}{c|}{ADE} & \multicolumn{3}{c|}{Cityscapes} & \multicolumn{2}{c|}{LVIS} & \multicolumn{2}{c|}{BDD}   & \multicolumn{2}{c|}{SCAN-20}  & \multicolumn{1}{c|}{SCAN-41}    & \multicolumn{1}{c}{SUN}    \\
 &SEG&DET&ITP & PQ & mask AP & box AP & {mIoU} & PQ & mask AP  & mIoU   & mask AP & box AP& PQ & mIoU   & PQ & mIoU  & mIoU & mIoU  \\ 
\hline
MSeg~(B)~\cite{lambert2020mseg} &\cmark&\xmark   &\xmark & \textcolor[rgb]{0.784,0.784,0.796}{33.7} & \textcolor[rgb]{0.784,0.784,0.796}{32.6} &$-$& 19.1 & 46.9 & \textcolor[rgb]{0.784,0.784,0.796}{24.8} & \textcolor[rgb]{0.784,0.784,0.796}{51.1}& $-$& $-$& $-$ & 44.9          &$-$& 33.4 & $-$&29.6        \\
MDETR~\cite{kamath2021mdetr} &\xmark&\cmark   &\cmark & $-$ & $-$& $-$ & $-$ &  $-$& $-$& $-$& $-$& \textcolor[rgb]{0.784,0.784,0.796}{24.2} & $-$& $-$& $-$& $-$& $-$& $-$ \\
LSeg+~(B)~\cite{li2022language}     &\cmark&\xmark   &\xmark & $-$ & $-$& $-$ & 18.0 &  $-$& $-$& $-$& $-$& $-$& $-$& $-$& $-$& $-$& $-$& $-$ \\
ZegFormer~(B)~\cite{hendricks2021decoupling} &\cmark&\xmark   &\xmark & $-$& $-$ & $-$ & 16.4 & $-$ &  $-$& $-$& $-$& $-$& $-$& $-$& $-$& $-$& $-$& $-$             \\
OpenSeg~(B)~\cite{gu2021open} &\cmark&\xmark   &\xmark & $-$ & $-$ & $-$ &21.1 & $-$& $-$& $-$&  $-$& $-$& $-$& $-$& $-$& $-$& $-$& $-$ \\
OpenSeg~(B)~\cite{gu2021open}           &\cmark&\xmark   &\cmark & $-$& $-$& $-$& 26.4 & $-$&  $-$& $-$& $-$& $-$& $-$& $-$& $-$& $-$& $-$& $-$ \\
MaskCLIP~(L)~\cite{ding2022open} &\cmark&\xmark   &\xmark & 15.1 & 6.0 & $-$ & 23.7 & $-$& $-$&  $-$& $-$& $-$& $-$& $-$& $-$& $-$& $-$& $-$\\ 
ODISE (H)~\cite{xu2023open}           &\cmark&\xmark   &\cmark & $23.5$ &$13.9$ & $-$ & $28.7$& $-$& $-$& $-$& $-$ & $-$& $-$& $-$& $-$& $-$& $-$& $-$ \\
GLIP (T)~\cite{li2022grounded}     &\xmark&\cmark   &\xmark & $-$ &$-$ & $-$ & $-$& $-$& $-$& $-$& $-$ & 18.5& $-$& $-$& $-$& $-$& $-$& $-$ \\
\hline
X-Decoder (T)~\cite{zou2022generalized} &\cmark&\xmark &\cmark &  18.8 &9.8 & $-$ & 25.0 & 37.2  &16.0 & 47.3 & $9.6$ & $-$ & 16.4&42.4  &30.7&37.8& 21.7 & 34.5 \\

\rowcolor[rgb]{0.961,0.961,0.961} {\ourmodel{} (T)} &\cmark&\cmark   &\xmark & \textbf{19.8} & \textbf{14.1}&\textbf{17.0} & 22.9 & \textbf{37.3} & \textbf{26.2} & 46.1& \textbf{19.4}&\textbf{21.8}& \textbf{17.2} &\textbf{44.8} &\textbf{39.7}& \textbf{45.1} & \textbf{25.2}       & \textbf{39.0} \\
\hline
X-Decoder (L)~\cite{zou2022generalized} &\cmark&\xmark   &\cmark     &  21.8 &13.1&$-$& 29.6 &38.1 &24.9&52.0  & $\star$ &   $-$ &17.8&47.2&39.5&49.5&29.7&43.0 \\
 
\rowcolor[rgb]{0.961,0.961,0.961} {\ourmodel{} (L)}  &\cmark&\cmark &\xmark & 19.7 & \textbf{15.0} & \textbf{17.7} & 23.4 & \textbf{41.4} & \textbf{33.2} & {47.8} & \textbf{21.0} & $\textbf{23.0}$ & \textbf{19.4} & \textbf{47.4} & {\textbf{42.2}} &$48.7$  & $27.4$&41.9
  \\
\bottomrule
\end{tabular}}
\vspace{0.2cm}
\end{table*}
\section{Experiment}
\subsection{Experimental Setup}
\noindent
\textbf{Datasets and Settings.}
In our experiments, we jointly pre-train on two types of data, including panoptic segmentation and object detection. For panoptic segmentation, we use 
COCO2017~\cite{lin2014microsoft} with segmentation annotations (around 110k images). For object detection, we use Objects365~\cite{shao2019objects365} (660k images for v1 and 1700k images for v2). We use Objects365v1 for training and ablating our tiny model and Objects365v2 only for training our large model. We evaluate our models on all tasks covered by pretraining, including semantic, instance, panoptic segmentation, and object detection. In particular, we benchmark on more than 60 datasets covering a wide range of domains on zero-shot segmentation and detection. 
\smallskip

\noindent
\textbf{Implementation Details.}
We build on Mask DINO~\cite{li2022mask} to implement our model. Mask DINO is a unified detection and segmentation framework which simultaneously predicts box and mask. We follow~\cite{li2022mask} to use 300 latent queries and nine decoder layers for thing categories in instance segmentation and add 100 panoptic queries for stuff categories. For the visual backbone, we adopt pretrained Swin-T/L~\cite{liu2021swin} by default. We also use Focal-T~\cite{yang2022focal} in our ablation studies following~\cite{zou2022generalized}. For the language backbone, we adopt the pretrained base model in UniCL~\cite{yang2022unified}. Particularly, our model only uses these pretrained backbones and does not use other image-text pairs or grounding data for pretraining~\cite{zou2022generalized,li2022grounded}. 
During pretraining, we set a minibatch for segmentation to $32$ and detection to $64$, and the image resolution is $1024\times 1024$ for both segmentation and detection. During fine-tuning, we use $512\times 1024$ for Cityscapes~\cite{cordts2015cityscapes} and $640\times 640$ for ADE20K~\cite{zhou2017scene} by default. Following the balanced sampling strategy in~\cite{yang2022unified, zou2022generalized}, the segmentation data are always sampled for a consistent number of epochs, regardless of the total number of detection data. We use AdamW~\cite{loshchilov2017decoupled} as the optimizer. We pre-train our model on the joint dataset for 30 epochs. The learning rate is set to $0.0001$, which is decayed at 0.9 and 0.95 fractions of the total number of steps by 10. \textit{Unless otherwise specified, we use online mask assistance during our pretraining by default}.

\subsection{Open-Vocabulary Benchmarking}
After pretraining our \ourmodel{} on COCO and Objects365, we evaluate it on a wide range of datasets in a zero-shot manner. Following~\cite{zou2022generalized}, we cover six commonly used segmentation datasets, including indoor scenes (ADE20K~\cite{zhou2017scene}), outdoor scenes (Cityscapes~\cite{cordts2015cityscapes}), and driving scenes (BDD100K~\cite{yu2018bdd100k}). In addition, we evaluate both segmentation and detection performance on LVIS~\cite{gupta2019lvis}. We report PQ, mask AP, and mIoU for panoptic, instance, and semantic segmentation, respectively, and use box AP for detection. The results are shown in Table~\ref{tab:open_seg}.

\begin{table*}
\caption{\textbf{Task-specific transfer} of \ourmodel{} to different segmentation and VL tasks. We directly evaluate the COCO performance without finetuning. Note: ``$-$" denotes the model does not have number reported or does not have the ability for the specific task.
$\star$ means it is the test set results. The results in the bracket are trained with 1280$\times$1280 image size.
Note that the results of GLIPv2 and X-Decoder on COCO are fine-tuned while those of \ourmodel{} are reported without fine-tuning. }
\label{tab:task_transfer}
\footnotesize  \setlength{\tabcolsep}{8.0pt}
\centering
\resizebox{0.999\linewidth}{!}{
\begin{tabular}{l|c|cccc|ccc|cccc} 
\specialrule{.12em}{.1em}{.1em}
\multirow{2}{*}{Method} & \multirow{2}{*}{Type} & \multicolumn{4}{c|}{\uline{ADE}} & \multicolumn{3}{c|}{\uline{Cityscapes}} & \multicolumn{4}{c}{\uline{COCO}} \\
 &  & PQ & mask AP & box AP & mIoU & PQ & mask AP & mIoU & PQ & mask AP & box AP & mIoU \\ 
\hline
Mask2Former (T)~\cite{cheng2022masked} & \multirow{7}{*}{Closed-set} & 39.7 & 26.4 & 28.8 & 47.7 & 63.9 & 39.1 & 80.5 & 53.2 & 43.3 & 46.1 & 63.2 \\
Mask2Former (B)~\cite{cheng2022masked} &  & $-$ & $-$ & $-$ & 53.9 & 66.1 & 42.8 & 82.7 & 56.4 & 46.3 & 49.5 & 67.1 \\
Mask2Former (L)~\cite{cheng2022masked} &  & 48.1 & 34.2 & 36.4 & 56.1 & 66.6 & 43.6 & 82.9 & 57.8 & {48.6} & 52.1 & 67.4 \\
OneFormer (L)~\cite{jain2022oneformer} &  & 48.6 & 35.9 & $-$ & 57.0 & 67.2 & 45.6 & ~83.0 & 57.9 & {48.9} & $-$ & 67.4 \\
MaskDINO (L)~\cite{li2022mask} &  & $-$ & $-$ & $-$ & $-$ & $-$ & $-$ & $-$ & 58.3 & {50.6} & 56.2 & 67.5 \\
Pano/SegFormer (B)~\cite{xie2021segformer} &  & $-$ & $-$ & $-$ & 51.0 & $-$ & $-$ & $-$ & 55.4 & $-$ & $-$ & $-$ \\
kMaX-DeepLab (L)~\cite{yu2022k} &  & 48.7 & $-$ & $-$ & 54.8 & $-$ & $-$ & $-$ & {58.1} & $-$ & $-$ & $-$ \\ 
\hline
GLIPv2 (T)~\cite{zhang2022glipv2} & \multirow{8}{*}{Open-vocabulary} & $-$ & $-$ & $-$ & $-$ & $-$ & $-$ & $-$ & $-$ & 42.0$^\star$ & $-$ & $-$ \\
GLIPv2 (B)~\cite{zhang2022glipv2} &  & $-$ & $-$ & $-$ & $-$ & $-$ & $-$ & $-$ & $-$ & 45.8$^\star$ & $-$ & $-$ \\
GLIPv2 (H)~\cite{zhang2022glipv2} &  & $-$ & $-$ &$-$& $-$ & $-$ & $-$ & & $-$ & 48.9$^\star$ & $-$ & $-$ \\
X-Decoder (T)~\cite{zou2022generalized} &  & 41.6 & 27.7 & 28.8 & 51.0 & 61.3 & 36.2& 78.7 & 52.6 &41.3 & 43.6 & 62.4 \\
{\cellcolor[rgb]{0.961,0.961,0.961}}\ourmodel{} (T) &  & {\cellcolor[rgb]{0.961,0.961,0.961}}\textbf{\textbf{47.2}} & {\cellcolor[rgb]{0.961,0.961,0.961}}\textbf{\textbf{35.1}} & {\cellcolor[rgb]{0.961,0.961,0.961}}\textbf{\textbf{39.4}} & {\cellcolor[rgb]{0.961,0.961,0.961}}\textbf{\textbf{52.2}} & {\cellcolor[rgb]{0.961,0.961,0.961}}\textbf{63.9} & {\cellcolor[rgb]{0.961,0.961,0.961}}\textbf{38.2}  & {\cellcolor[rgb]{0.961,0.961,0.961}}\textbf{80.3} & 
{\cellcolor[rgb]{0.961,0.961,0.961}}
{\textbf{55.4}} & {\cellcolor[rgb]{0.961,0.961,0.961}}\textbf{47.6} & {\cellcolor[rgb]{0.961,0.961,0.961}}\textbf{52.0} & {\cellcolor[rgb]{0.961,0.961,0.961}}\textbf{64.0} \\
X-Decoder (L)~\cite{zou2022generalized} &  & 49.6 & 35.8 & $-$ & 58.1 & 65.6 & 42.2 & 81.7 & 56.9 & 46.7 & $-$ & 67.5 \\

{\cellcolor[rgb]{0.961,0.961,0.961}}\ourmodel{} (L) &  & {\cellcolor[rgb]{0.961,0.961,0.961}}\textbf{53.1~(53.7)} & {\cellcolor[rgb]{0.961,0.961,0.961}}\textbf{42.0~(42.6)} & {\cellcolor[rgb]{0.961,0.961,0.961}}\textbf{46.4~(46.9)} & {\cellcolor[rgb]{0.961,0.961,0.961}}\textbf{58.6~(58.4)} & {\cellcolor[rgb]{0.961,0.961,0.961}\textbf{69.2}}& {\cellcolor[rgb]{0.961,0.961,0.961}\textbf{49.3}}& {\cellcolor[rgb]{0.961,0.961,0.961}\textbf{84.5}}& {\cellcolor[rgb]{0.961,0.961,0.961}\textbf{59.5}}  & {\cellcolor[rgb]{0.961,0.961,0.961}\textbf{53.2}} & {\cellcolor[rgb]{0.961,0.961,0.961}\textbf{58.2}}& {\cellcolor[rgb]{0.961,0.961,0.961}\textbf{68.6}} \\
\specialrule{.12em}{.1em}{.1em}
\end{tabular}
}
\end{table*}

\begin{table*}[t]
\vspace{4pt}
\caption{\textbf{One suit of weights} on the SeginW benchmark in a zero-shot manner.}
\label{tab:seginw}
\tiny  \setlength{\tabcolsep}{1.3pt}
\centering
\tablestyle{3pt}{0.95}
\addtolength{\extrarowheight}{\belowrulesep}
\footnotesize
\resizebox{0.99\linewidth}{!}{
\begin{tabu}{l|cc|ccccccccccccccccccccccccc} 
\specialrule{.12em}{.1em}{.1em}
Model & Med. & Avg & \begin{tabular}[c]{@{}c@{}}Air-\\Par.\end{tabular} & Bottles & \begin{tabular}[c]{@{}c@{}}Br.\\Tum.\end{tabular} & Chicken & Cows & \begin{tabular}[c]{@{}c@{}}Ele.-\\Sha.\end{tabular} & Eleph. & Fruits & Gar. & \begin{tabular}[c]{@{}c@{}}Gin.-\\Gar.\end{tabular} & Hand & \begin{tabular}[c]{@{}c@{}}Hand-\\Metal\end{tabular} & \begin{tabular}[c]{@{}c@{}}House-\\Parts\end{tabular} & \begin{tabular}[c]{@{}c@{}}HH.-\\Items\end{tabular} & \begin{tabular}[c]{@{}c@{}}Nut.-\\Squi.\end{tabular} & Phones & Poles & Puppies & Rail & \begin{tabular}[c]{@{}c@{}}Sal.-\\Fil.\end{tabular} & Stra. & Tablets & Toolkits & Trash & W.M \\ 
\hline
X-Decoder (T)~\cite{zou2022generalized} & 15.2 & 22.7 & 10.5 & 19.0 & 1.1 & 12.0 & 12.0 & 1.2 & 65.6 & 66.5 & 28.7 & 7.9 & 0.6 & 22.4 & 5.5 & 50.6 & 62.1 & 29.9 & 3.6 & 48.9 & 0.7 & 15.0 & 41.6 & 15.2 & 9.5 & 19.3 & 16.2 \\
\ourmodel{} (T) & 21.5 & \textbf{33.9} & 12.2 & 27.4 & 5.0 & 68.7 & 21.5 & 0.3 & 73.3 & 72.9 & 7.3 &  6.2 & 92.4  & 62.3 & 0.5 & 55.0 & 63.6 & 2.4 & 4.6 & 63.8 & 5.4 & 15.6 & 85.3& 32.0 & 4.8 & 14.5 & 51.0 \\
\hline
X-Decoder (L)~\cite{zou2022generalized} & 22.3 & 32.3 & 13.1 & 42.1 & 2.2 & 8.6 & 44.9 & 7.5 & 66.0 & 79.2 & 33.0 & 11.6 & 75.9 & 42.1 & 7.0 & 53.0 & 68.4 & 15.6 & 20.1 & 59.0 & 2.3 & 19.0 & 67.1 & 22.5 & 9.9 & 22.3 & 13.8 \\ 
\ourmodel{} (L) & 38.7 & \textbf{36.1} & 13.0 & 39.7 & 2.1 & 82.9 & 40.9 & 4.7 & 72.9 & 76.4 & 16.9 & 13.6 & 92.7 & 38.7 & 1.8 & 50.0 & 40.0 & 7.6 & 4.6 & 74.6 & 1.8 & 15.0 & 82.8 & 47.4 & 15.4 & 15.3 & 52.3 \\
\specialrule{.12em}{.1em}{.1em}
\end{tabu}}
\vspace{-.2cm}
\end{table*}

We first compare with previous works on segmentation tasks. Overall, our model achieves significantly better performance on instance segmentation and comparable performance for panoptic and semantic segmentation. Compared with state-of-the-art methods ODISE~\cite{xu2023open} and X-Decoder~\cite{zou2022generalized}, \ourmodel{} achieves \textbf{1.1} and \textbf{1.9} mask AP improvements on ADE20K, respectively. This gap is even larger on Cityscapes and LVIS. Our \ourmodel{} outperforms X-Decoder by \textbf{10.2} and \textbf{8.3} mask AP with a tiny and large model on Cityscapes, respectively. On LVIS, we evaluate the mask AP with the released X-Decoder tiny model and the comparison shows \textbf{9.8} mask AP improvement with our \ourmodel{} tiny model. These results indicate that the proposed joint learning method can effectively transfer the instance-level knowledge in detection data for instance segmentation. Compared with instance segmentation, both panoptic and semantic segmentation requires the segmentation of background stuff, which is fully absent in the detection data. Despite that, our \ourmodel{} still outperforms X-Decoder for panoptic segmentation on 3 out of 4 datasets (except for ADE20K), and achieves comparable semantic segmentation performance. The results on three segmentation tasks suggest that detection data significantly benefit the instance-level understanding while image-text pairs mainly augment the semantic understanding for semantic segmentation.
In addition to segmentation, \ourmodel{} also produces reasonably good detection performance. Compared with GLIP, our \ourmodel{}~(T) outperforms GLIP~(T) (setting A) for zero-shot detection on LVIS (\textbf{21.8}~\textit{v.s.}~18.5), where both only use Objects365 as the pretraining detection dataset. \emph{At last, we highlight that our model is the first one that can be pretrained with segmentation and detection data jointly and perform zero-shot transfer to both tasks.}
\begin{table}[t]
    \centering
    \caption{\textbf{One suit of weights} on ODinW benchmark. Average and median AP across 35 datasets are reported for simplicity.}
    \begin{adjustbox}{width=0.45\textwidth,center}
    \begin{tabular}{l|c|c|ccc|ccc}
    \toprule
        Model  & Pretrain Data & Average & Median   \\
        \hline
        MDETR~\cite{kamath2021mdetr} &GOLDG, REFC &10.7&3.0\\
        GLIP-T~\cite{li2022grounded}    & Object365 &11.4&1.6\\
        
        \ourmodel{} (T) (ours)    &Object365, COCO&{14.2}&{3.1}   \\

        \ourmodel{} (L) (ours)  &Object365, COCO& \textbf{15.2}&\textbf{5.0} \\
        
        \bottomrule
    \end{tabular}
    \end{adjustbox}
    \centering

    \label{tab:panoptic}
    \vspace{-.0cm}
\end{table}

\subsection{Direct and Task-Specific Transfer}
After pretraining, our model can be directly transferred to downstream segmentation and detection tasks. In Table~\ref{tab:task_transfer}, we compare our model with both closed-set and open-vocabulary methods. Remarkably, our model achieves SOTA performance \textbf{59.5} PQ for COCO panoptic segmentation \textit{without any further fine-tuning}. After data-specific fine-tuning, \ourmodel{} establishes a new SOTA on ADE20K panoptic~(\textbf{53.7} PQ) and instance segmentation~(\textbf{42.6} AP) when trained with $1280\times 1280$ image size. In addition, we also achieve a new SOTA on Cityscapes instance segmentation~(\textbf{48.5} AP). These results indicate the jointly pretrained open-vocabulary model can also be well transferred to closed-set detection and segmentation.

\subsection{Segmentation and Detection in the Wild}

To investigate the generalization ability of \ourmodel{} for segmentation, we evaluate our model on more domain-specific datasets. We conduct a zero-shot evaluation of our model on the Segmentation in the Wild (SeginW) benchmark~\cite{zou2022generalized}, which includes 25 datasets. As this benchmark focuses on instance segmentation, we report the average and median mAP of all the datasets following the common practice. The results in Table~\ref{tab:seginw} indicate that combining detection supervision significantly improves the segmentation performance by more than \textbf{10} AP under the same setting.

To further study the object detection ability of \ourmodel{}, we follow GLIP~\cite{li2022grounded} to evaluate  detection performance on Object Detection in the Wild (ODinW) benchmark. It collects over 35 datasets and is closer to real-world scenarios. We report the average and median mAP of all 35 datasets. With the jointly pretrained model weights, we directly evaluate this challenging benchmark in a zero-shot manner.  Under the same setting that only uses Objects365 as the detection data for training, our tiny model outperforms GLIP-T (setting A) by $2.8$ AP in average.


\begin{table}
\centering
\tablestyle{3pt}{0.95}
\addtolength{\extrarowheight}{\aboverulesep}
\caption{The performance of \ourmodel{} on close-set segmentation. We train \ourmodel{} with only COCO ('w/o DET') data to compare with the top-performed close-set segmentation model MaskDINO. We also train \ourmodel{} on COCO and Object365 together to verify the ability to utilize detection data. $^*$ none of our designed components are used.}
\footnotesize
\resizebox{1.\linewidth}{!}{
\begin{tabular}{l|cccc|cccc} 
\toprule
\multirow{2}{*}{Method}                        
&\multicolumn{4}{c|}{\uline{ADE}}   & \multicolumn{4}{c}{\uline{COCO}} \\
& PQ  & mask AP  & box AP & {mIoU}    & PQ & mask AP & box AP & mIoU \\ 
\cline{1-9}
MaskDINO-ResNet-50$^*$~\cite{li2022mask} & $-$&$-$&$-$&$-$ &  53.0	&44.3	&48.8	&60.6                             \\
\ourmodel{}-ResNet-50$^*$ w/o DET & $-$&$-$&$-$&$-$ &  52.7	&44.0	&47.9	&60.2                             \\
\hline
\ourmodel{}-SwinT$^*$ w/o DET &16.0&9.9&10.7&18.6&  54.0 & 45.6 &49.0&62.1                                             \\
\ourmodel{}-SwinT  &\textbf{19.8}& \textbf{14.1} &\textbf{17.0} &\textbf{22.9} &  \textbf{55.2}&\textbf{47.3}&\textbf{51.9}& \textbf{63.7}                                          \\
\hline
\ourmodel{}-FocalT$^*$ w/o DET &15.0&8.7&9.2&17.5&  54.0 & 45.1 &47.5&62.5                                             \\
\ourmodel{}-FocalT &\textbf{18.5}& \textbf{13.4} &\textbf{15.6} &\textbf{21.7}&  \textbf{55.4}&\textbf{47.3}&\textbf{51.0}& \textbf{63.4}                                           \\
\bottomrule
\end{tabular}
}
\label{tab:ablation:coco}
\end{table}

\begin{table}
\centering
\tablestyle{3pt}{0.95}
\addtolength{\extrarowheight}{\aboverulesep}
\vspace{3pt}
\caption{\textbf{Ablation} of the effectiveness of pseudo annotations in offline mask assistance for our open-vocabulary model. We evaluate the model performance on ADE20K and COCO. "-anno" denotes with annotations and "w/o anno" denotes without annotations.}
\begin{adjustbox}{width=0.47\textwidth,center}
\begin{tabular}{l|cccc|cccc} 
\toprule
\multirow{2}{*}{Method}                        
&\multicolumn{4}{c|}{\uline{ADE}}   & \multicolumn{4}{c}{\uline{COCO}} \\
& PQ  & mask AP & box AP & {mIoU}    & PQ & mask AP& box AP  & mIoU \\ 
\cline{1-9}

\ourmodel{}-FocalT w/o anno  &  18.4 & 12.6 &14.5&21.4&55.1&46.8& 50.3& 62.7                                           \\
\ourmodel{}-FocalT-anno  &  18.2 & \textbf{14.1}\fontsize{8.0pt}{\baselineskip}\selectfont{(\textbf{+1.5})}&\textbf{16.2}\fontsize{8.0pt}{\baselineskip}\selectfont{(\textbf{+1.7})}&21.4&55.2&\textbf{47.5}\fontsize{8.0pt}{\baselineskip}\selectfont{(\textbf{+0.7})}& \textbf{51.3}\fontsize{8.0pt}{\baselineskip}\selectfont{(\textbf{+1.0})} &63.4                                          \\
\ourmodel{}-SwinT w/o anno  &  19.8& 14.1 &17.0&22.9 &  55.2&47.3&51.9& 63.7                                            \\
\ourmodel{}-SwinT-anno  &  \textbf{20.4}\fontsize{8.0pt}{\baselineskip}\selectfont{(\textbf{+0.6})}& \textbf{14.8}\fontsize{8.0pt}{\baselineskip}\selectfont{(\textbf{+0.7})}&\textbf{17.1}&\textbf{24.0}\fontsize{8.0pt}{\baselineskip}\selectfont{(\textbf{+1.1})}&55.5&\textbf{47.9}\fontsize{8.0pt}{\baselineskip}\selectfont{(\textbf{+0.6})}& \textbf{52.4}\fontsize{8.0pt}{\baselineskip}\selectfont{(\textbf{+0.5})} &63.9                                          \\

\bottomrule
\end{tabular}
\end{adjustbox}
\label{tab:ablation:offline anno}
\end{table}
\subsection{Ablation}
\noindent
\textbf{Ablation on our basic components.} We first ablate on our basic model to verify whether each basic component works as expected. We evaluate our model on the COCO closed-set panoptic segmentation task. The first two rows in Table~\ref{tab:ablation:coco} show that our model can achieve a comparable closed-set panoptic segmentation performance as MaskDINO using the same ResNet-50 backbone. It is also shown in the last four rows that our framework can significantly improve the segmentation performance by combining tasks and utilizing detection data across different backbones.
\\
\textbf{Ablation on the offline mask supervision.}
As we discussed earlier, the proposed conditioned mask decoding can be used either in online or offline manner. Here, to investigate the effectiveness of offline setting, we generate pseudo annotations with our large models and then use them to tune the tiny models. Concretely, we first generate pseudo masks on Objects365 conditioned on boxes with our \ourmodel{}~(L) model. Given the masks, we conduct experiments on two models, including the open-vocabulary model (\ourmodel{}) and the closed-set model (MaskDINO) by using the pseudo masks for supervision. As shown in Table~\ref{tab:ablation:offline anno}, when evaluating our models, we find that the mask AP and box AP are significantly improved.
When conducting experiments on MaskDINO, we perform object detection for the setting without pseudo annotation and instance segmentation for the setting with pseudo annotations during pretraining. After pretraining, we do a zero-shot evaluation on COCO. We also use the pretrained model for fine-tuning. The results in Table~\ref{tab:ablation:offline anno closedset} indicate that in both settings, pseudo-annotations improve performance. It is also shown that the box AP can also be improved accordingly with extra mask annotations. In addition, we are the first to report zero-shot segmentation performance on COCO which is comparable with the full-shot performance of other methods.
\\
\noindent
\textbf{Ablation on decoupled decoding and online mask assistance.} In Table~\ref{tab:ablation:query split}, we conduct experiments to show the effectiveness of our proposed components by removing them one at a time. In the second row, after removing online mask assistance, the instance mask and box performance on the open-segmentation dataset ADE20K drops by 0.6 AP and 0.5 AP, respectively. In the third row, when we remove the decoupled decoding, the performance of the mask and box is further impacted by a large margin (-1.1 AP and -2.3 AP for ADE mask and box). Both results suggest that the proposed techniques help to mitigate the gap between segmentation and detection data.


\begin{table}
\centering
\tablestyle{3pt}{0.95}
\addtolength{\extrarowheight}{\aboverulesep}
\caption{\textbf{Ablation} of the effectiveness of pseudo annotations in offline mask assistance for close-set segmentation models. We conduct experiments with maskDINO pre-trained on Object365 with Mask DINO. The task is instance segmentation and object detection for the settings with or without pseudo annotations, respectively. "O365+anno" denotes Objects365 data with pseudo-annotations of masks.}
\begin{adjustbox}{width=0.47\textwidth,center}
\begin{tabular}{l|c|cc|cc} 
\toprule
\multirow{2}{*}{Method}  &\multirow{2}{*}{Pre-training Dataset}                        
&\multicolumn{2}{c|}{\uline{COCO zeroshot}}   & \multicolumn{2}{c}{\uline{COCO finetune}} \\
& & mask AP & box AP    & mask AP& box AP  \\ 
\cline{1-6}
GLIP A(Swin-T)~\cite{li2022grounded}&O365  &  $-$ & 42.9 & $-$ &52.9                                          \\
GLIP B(Swin-T)~\cite{li2022grounded}&O365  &  $-$ & 44.9 & $-$ &53.8                                         \\
MaskDINO-R50 & O365 &  $-$ & 42.4 &46.7&52.0                                   \\
MaskDINO-R50&O365+anno &  \textbf{41.4} & \textbf{44.3}\fontsize{8.0pt}{\baselineskip}\selectfont{(\textbf{+1.9})}&\textbf{48.5}\fontsize{8.0pt}{\baselineskip}\selectfont{(\textbf{+1.8})}&\textbf{54.3}\fontsize{8.0pt}{\baselineskip}\selectfont{(\textbf{+2.3})}                                        \\

\bottomrule
\end{tabular}
\end{adjustbox}
\label{tab:ablation:offline anno closedset}
\end{table}

\begin{table}
\centering
\addtolength{\extrarowheight}{\aboverulesep}
\caption{\textbf{Ablation} of the decoupled decoding and the online mask assistance. We keep other settings the same and only remove one design at a time. For the setting without decoupled decoding, we let foreground and background queries be selected together from image features.  }
\begin{adjustbox}{width=0.47\textwidth,center}
\begin{tabular}{l|cccc|cccc} 
\toprule
\multirow{2}{*}{Method}                        
&\multicolumn{4}{c|}{\uline{ADE}}   & \multicolumn{4}{c}{\uline{COCO}} \\
& PQ  & mask AP  & box AP & {mIoU}    & PQ & mask AP & box AP & mIoU \\ 
\cline{1-9}
\ourmodel{}-SwinT  & 19.8& 14.2 &17.2 &23.1&55.2&47.3&51.9&63.7 \\
$-$ online mask assistance  &  19.4 & \textbf{13.6}\fontsize{8.0pt}{\baselineskip}\selectfont{(\textbf{-0.6})} &\textbf{16.7}\fontsize{8.0pt}{\baselineskip}\selectfont{(\textbf{-0.5})}&22.9&55.0&47.5&52.3& 63.9 \\
$-$ decoupled decoding &  19.0 & \textbf{13.1}\fontsize{8.0pt}{\baselineskip}\selectfont{(\textbf{-1.1})} &\textbf{14.9}\fontsize{8.0pt}{\baselineskip}\selectfont{(\textbf{-2.3})}&22.4& 55.0&\textbf{46.4}\fontsize{8.0pt}{\baselineskip}\selectfont{(\textbf{-0.9})}&\textbf{49.5}\fontsize{8.0pt}{\baselineskip}\selectfont{(\textbf{-2.4})}& 63.9                                             \\


\bottomrule
\end{tabular}
\end{adjustbox}
\label{tab:ablation:query split}
\vspace{-0.3cm}
\end{table}


    
    

\section{Conclusion}
We have presented \ourmodel{}, a simple {open}-vocabulary {se}gm{e}ntation and {d}etection framework, which jointly learns from different segmentation and detection datasets with a single model. To bridge the task gap between foreground objects and background stuff, we propose a decoupled decoding method with language-guided foreground query selection. We also jointly train a conditioned mask decoding task, which provides an interactive segmentation interface during inference and helps bridge data gap for detection data during training. The result indicates our unified model significantly improves open-segmentation while keeping a reasonable detection performance. The jointly pre-trained model can also be seamlessly transferred to improve close-vocabulary performance. 
\\
\textbf{Limitations}. In this work, we aim at exploring the potential of training an open-vocabulary model for both segmentation and detection. \ourmodel{} does not utilize either referring/grounding data or large-scale image-text pairs to further enrich our training data and semantic coverage. We leave a grander joint training to future work. 


{\small
\bibliographystyle{ieee_fullname}
\bibliography{egbib}
}

\end{document}


\title{---Supplementary Materials---\\  A Simple Framework for Open-Vocabulary Segmentation and Detection}

\author{First Author\\
Institution1\\
Institution1 address\\
{\tt\small firstauthor@i1.org}
\and
Second Author\\
Institution2\\
First line of institution2 address\\
{\tt\small secondauthor@i2.org}
}

\maketitle
\ificcvfinal\thispagestyle{empty}\fi

\appendix
\section*{Overview}
This supplementary material presents more details and additional results not included in the main paper due to page limitation. The list of items included are:

\begin{itemize}
    \item Correction of typos in Sec.~\ref{typo}.
    \item More experimental results in Sec.~\ref{exp}.
    \item Visualization of the predictions of \ourmodel{} in Sec.~\ref{vis}.
    \item More implementation details in Sec.~\ref{imp}.
\end{itemize}
\section{Typo Correction}
\label{typo}
 There are two typos in the paper. We are sorry if they affect your reading. We will fix them in the next version.\\
 \begin{enumerate}
 \item There is a typo of duplicate sentences ``Mask2Former and MaskDINO treat ... handling foreground and background." in ``Comparison with previous works " in section 3.2. 
 \item There is a typo in Row 3 in Table 8. The correct version should be Table~\ref{tab:ablation:offline anno closedset}.
 \end{enumerate}
 \setcounter{table}{9}

\begin{table}
\centering
\tablestyle{3pt}{0.95}
\addtolength{\extrarowheight}{\aboverulesep}
\caption{\textbf{Ablation} of the effectiveness of pseudo annotations in offline mask assistance for close-set segmentation models. We conduct experiments with maskDINO pre-trained on Object365 with Mask DINO. The task is instance segmentation and object detection for the settings with or without pseudo annotations, respectively. "O365+anno" denotes Objects365 data with pseudo-annotations of masks.}
\begin{adjustbox}{width=0.47\textwidth,center}
\begin{tabular}{l|c|cc|cc} 
\toprule
\multirow{2}{*}{Method}  &\multirow{2}{*}{Pre-training Dataset}                        
&\multicolumn{2}{c|}{\uline{COCO zeroshot}}   & \multicolumn{2}{c}{\uline{COCO finetune}} \\
& & mask AP & box AP    & mask AP& box AP  \\ 
\cline{1-6}
GLIP A(Swin-T)~\cite{li2022grounded}&O365  &  $-$ & 42.9 & $-$ &52.9                                          \\
GLIP B(Swin-T)~\cite{li2022grounded}&O365  &  $-$ & 44.9 & $-$ &53.8                                         \\
MaskDINO-R50 & O365 &  $-$ & 42.4 &46.7&52.0                                   \\
MaskDINO-R50&O365+anno &  \textbf{41.4} & \textbf{44.3}\fontsize{8.0pt}{\baselineskip}\selectfont{(\textbf{+1.9})}&\textbf{48.5}\fontsize{8.0pt}{\baselineskip}\selectfont{(\textbf{+1.8})}&\textbf{54.3}\fontsize{8.0pt}{\baselineskip}\selectfont{(\textbf{+2.3})}                                        \\

\bottomrule
\end{tabular}
\end{adjustbox}
\label{tab:ablation:offline anno closedset}
\end{table}

\section{More Experimental Results}
\label{exp}
\subsection{SwinL 4-scale results}
Our Swin-L results in Table 2 adopts 5 scales of image features. In order to compare with other methods more thoroughly, we show the performance of our model with Swin-L and 4 scales of image features in Table~\ref{tab_supp:open_seg_swinl}.

\begin{table}
\centering
\tablestyle{3pt}{0.95}
\addtolength{\extrarowheight}{\belowrulesep}
\caption{\textbf{One suit of weights} for open-vocabulary segmentation on multiple datasets in a zero-shot manner. Our model is pre-trained on COCO and Objects365 data. 'SEG' indicates segmentation data (COCO), 'DET' indicates detection data (Objects365), and ITP indicates image-text pairs/referring/captioning data. The values in gray are supervised results. $\star$ X-Decoder (L) is not open-source, so we cannot evaluate its performance on LVIS.}
\label{tab_supp:open_seg_swinl}
\footnotesize
\resizebox{0.99\linewidth}{!}{\begin{tabular}{l|ccc|cccc} 
\toprule
\multirow{2}{*}{Method} & \multicolumn{3}{c|}{Training Data} & \multicolumn{4}{c|}{ADE}   \\
 &SEG&DET&ITP & PQ & mask AP & box AP & {mIoU}  \\ 
\hline
X-Decoder (L)~\cite{zou2022generalized} &\cmark&\xmark   &\cmark     &  21.8 &13.1&$-$& 29.6 \\
\rowcolor[rgb]{0.961,0.961,0.961} {\ourmodel{} (L)}  &\cmark&\cmark &\xmark & 20.3 & \textbf{15.0} & \textbf{18.3} & 23.6 
  \\
\bottomrule
\end{tabular}}
\vspace{0.2cm}
\end{table}
\subsection{Offline Mask Guidance Ablation with SwinT}
In order to be coherent with Table 2, we also show the ablation of offline mask assistance in Table~\ref{tab_supp:ablation: offline anno swint}, which verifies the effectiveness of our pseudo-annotations.

\begin{table}
\centering
\tablestyle{3pt}{0.95}
\addtolength{\extrarowheight}{\aboverulesep}
\vspace{3pt}
\caption{\textbf{Ablation} of the effectiveness of pseudo annotations in offline mask assistance for our open-vocabulary model. We evaluate the model performance on ADE20K and COCO. "-anno" denotes with annotations and "w/o anno" denotes without annotations.}
\begin{adjustbox}{width=0.47\textwidth,center}
\begin{tabular}{l|cccc|cccc} 
\toprule
\multirow{2}{*}{Method}                        
&\multicolumn{4}{c|}{\uline{ADE}}   & \multicolumn{4}{c}{\uline{COCO}} \\
& PQ  & mask AP & box AP & {mIoU}    & PQ & mask AP& box AP  & mIoU \\ 
\cline{1-9}

\ourmodel{}-SwinT w/o anno  &  19.8& 14.1 &17.0&22.9 &  55.2&47.3&51.9& 63.7                                            \\
\ourmodel{}-SwinT-anno  &  \textbf{20.4}\fontsize{8.0pt}{\baselineskip}\selectfont{(\textbf{+0.6})}& \textbf{14.8}\fontsize{8.0pt}{\baselineskip}\selectfont{(\textbf{+0.7})}&\textbf{17.1}&\textbf{24.0}\fontsize{8.0pt}{\baselineskip}\selectfont{(\textbf{+1.1})}&55.5&\textbf{47.9}\fontsize{8.0pt}{\baselineskip}\selectfont{(\textbf{+0.6})}& \textbf{52.4}\fontsize{8.0pt}{\baselineskip}\selectfont{(\textbf{+0.5})} &63.9                                          \\

\bottomrule
\end{tabular}
\end{adjustbox}
\label{tab_supp:ablation: offline anno swint}
\end{table}
\section{Visualization}
In this section, we show a visualization of \ourmodel{} for open-vocabulary segmentation on ADE20K and Objects365 we also show the conditioned segmentation ability of \ourmodel{}. Note that all experiments here utilize the model jointly trained on COCO panoptic segmentation and Objects365 detection without fine-tuning.
\label{vis}
\subsection{Three open-vocabulary segmentation tasks on ADE20K}
\begin{figure*}[t]
    \centering
    \includegraphics[width=0.95\linewidth]
    {iccv2023AuthorKit/resources/imgs/supp_basic_seg.pdf}
    \caption{Visualizations of \ourmodel{} for open-vocabulary instance segmentation and detection, panoptic and semantic segmentation.}
    \label{fig:supp basic seg}
    \vspace{-3pt}
\end{figure*}
In Fig.~\ref{fig:supp basic seg}, we show a visualization of \ourmodel{} for open-vocabulary instance segmentation and detection, panoptic and semantic segmentation on ADE20K dataset without finetuning.
\subsection{Segmentation in Objects365 Categories}
\begin{figure*}[t]
    \centering
    \includegraphics[width=0.95\linewidth]
    {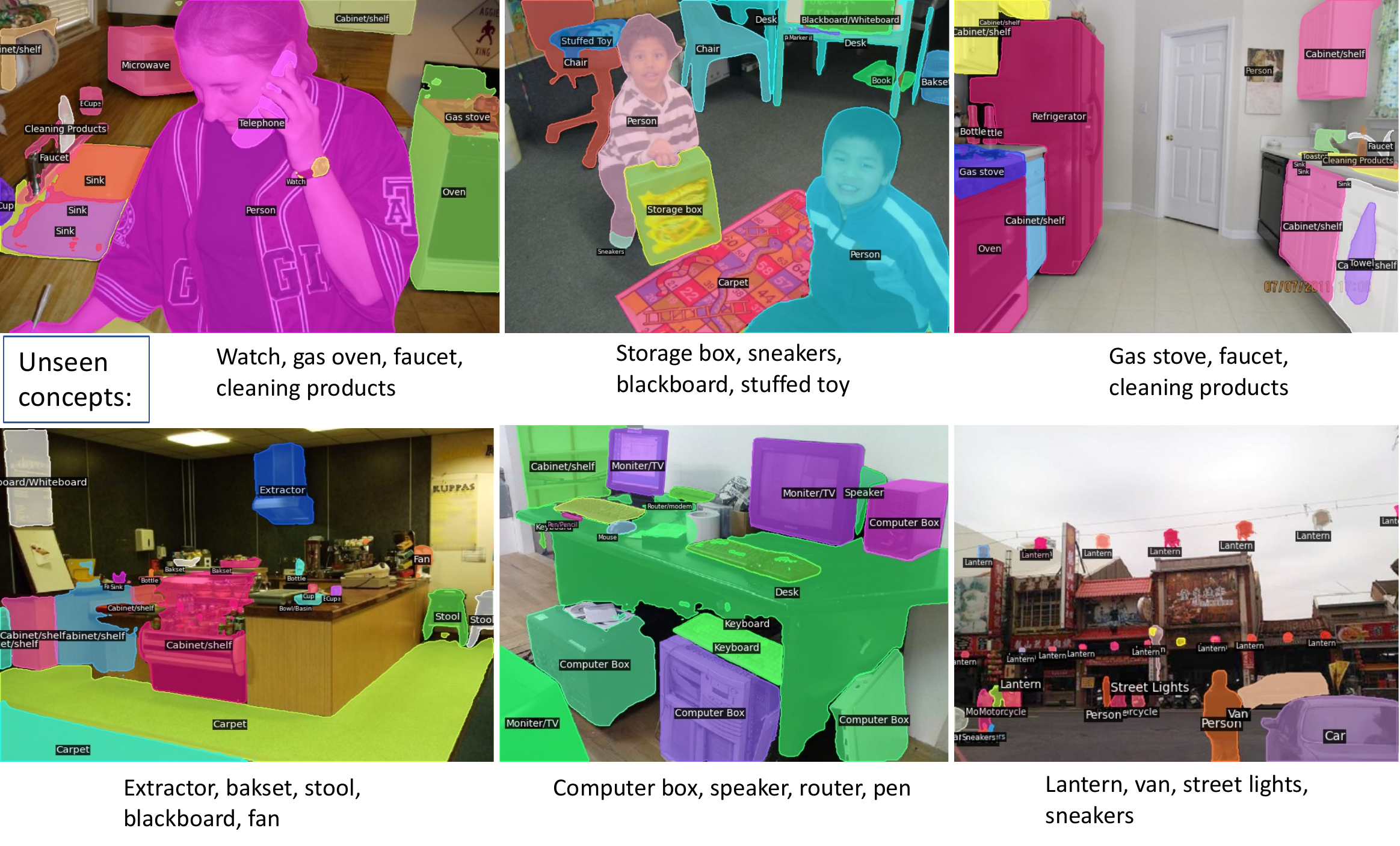}
    \caption{Visualizations of \ourmodel{} for open-vocabulary instance segmentation and detection, panoptic and semantic segmentation.}
    \label{fig:supp o365 unseen}
    \vspace{-3pt}
\end{figure*}
In Fig.~\ref{fig:supp o365 unseen}, we show the instance segmentation on Objects365 where unseen concepts are listed under each image. Unseen concepts are the categories that do not exist in COCO, which means they are not trained with segmentation annotations. Our model can segment instances from the unseen categories well although it is only trained with detection task on these categories.
\subsection{Conditioned Segmentation}
\begin{figure*}[t]
    \centering
    \includegraphics[width=0.95\linewidth]
    {iccv2023AuthorKit/resources/imgs/supp_cond_ade.pdf}
    \caption{Visualizations of \ourmodel{} for open-vocabulary instance segmentation and detection, panoptic and semantic segmentation.}
    \label{fig:supp cond ade}
    \vspace{-3pt}
\end{figure*}
In Fig.~\ref{fig:supp cond ade}, we show the conditioned segmentation ability of \ourmodel{}. When we give different conditioned boxes and text, we can obtain the corresponding masks.
\section{More Implementation Details}
\label{imp}
\begin{figure}[t]
    \centering
    \includegraphics[width=0.98\linewidth]
    {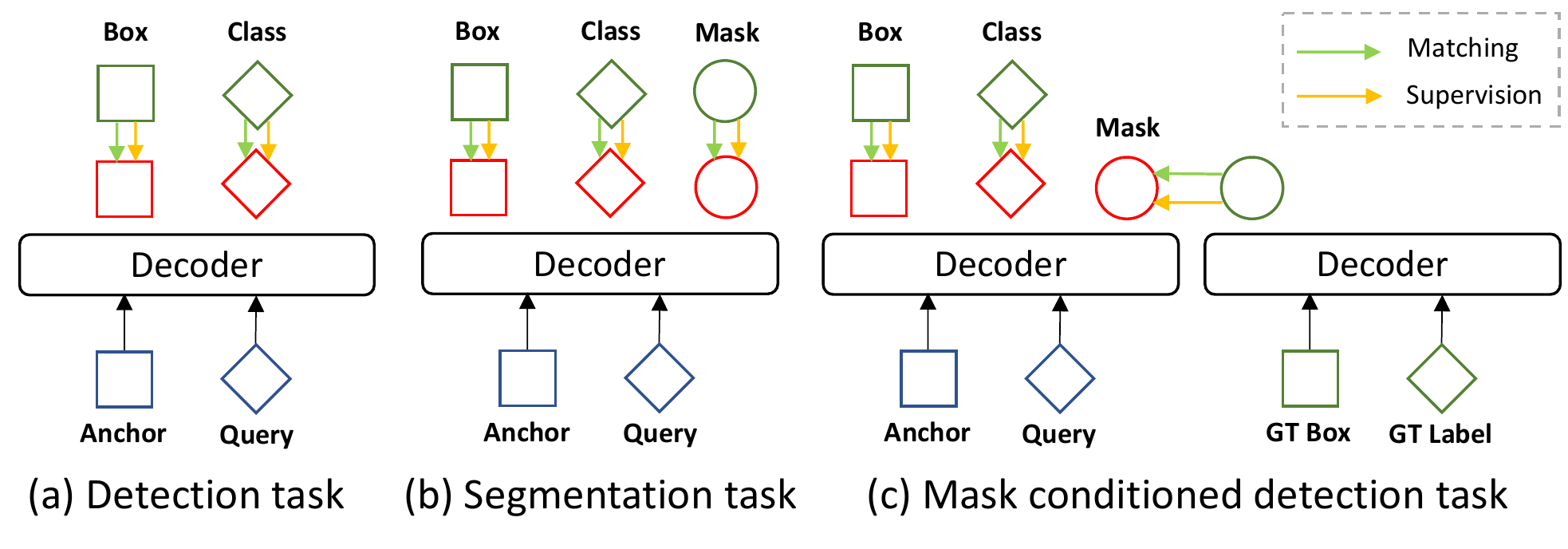}
    \caption{(a)(b) Standard detection and segmentation. They differ in that the segmentation task has mask supervision and matching. (c) In our online training, the detection task is assisted by conditioned-generated masks when matching. }
    \label{fig:cond matching}
\end{figure}
\textbf{Online Conditioned Mask as the Guidance. }
\noindent 
Given that our model can generate reasonably good masks with GT boxes as the condition, we seek to 
use them to better align detection with segmentation. A straightforward way is directly using masks for supervision on the fly. This requires high mask quality during the whole training phase, which however is not true, especially in the early training stage. 
Therefore, we alternatively use the generated mask as the additional guidance to find the matched foreground queries with the GT concept and box in detection data. 
As shown in Fig.~\ref{fig:cond matching}~(a), detection during training fully ignores the predicted mask quality when finding the matched foreground queries, which is different from segmentation in Fig.~\ref{fig:cond matching}~(b). This ignorance may lead to a biased matching toward box quality for our model which needs to produce high-quality boxes and masks simultaneously. Therefore, we use the generated masks from GT boxes and labels as the guidance for better label assignment, as shown in Fig.~\ref{fig:cond matching}~(c). 
\\
\textbf{Offline Conditioned Mask as the Supervision. } Another way to better use the generated mask is to train two models. For example, we can train a large conditioned mask decoding model first and then use it to generate pseudo masks for all the detection data. Afterward, we can train the second model on the annotated detection dataset with mask supervision. However, though our model can generate fairly good masks for novel categories, these masks still has inferior quality compared to the trained categories. Considering that 85 categories in Objects365~\cite{shao2019objects365} are in the COCO foreground category set, we treat the generated annotations on these categories as \textit{golden annotations} because they are trained with mask annotations on COCO. Other annotations are \textit{coarse annotations}. We adopt different strategies for the two types of annotations, where \textit{golden annotations} are used for mask supervision while \textit{coarse annotations} are only used for matching (similar to our online guidance). 
{\small
\bibliographystyle{ieee_fullname}
\bibliography{egbib}
}